\documentclass[pmlr]{jmlr}

\usepackage{longtable}

 \usepackage{booktabs}

\usepackage[load-configurations=version-1]{siunitx} 

 \usepackage{cleveref}
 \usepackage{color,soul}

\makeatletter
\def\set@curr@file#1{\def\@curr@file{#1}} 
\makeatother

\theorembodyfont{\upshape}
\theoremheaderfont{\scshape}
\theorempostheader{:}
\theoremsep{\newline}

\jmlrvolume{219}
\jmlryear{2023}
\jmlrworkshop{Machine Learning for Healthcare}

\title[When More is Less: Incorporating Additional Datasets Can Hurt Performance]{When More is Less: Incorporating Additional Datasets Can Hurt Performance By Introducing Spurious Correlations}

\author{\Name{Rhys Compton}
       \Email{rhys.compton@nyu.edu}\\ 
       \Name{Lily Zhang}
       \Email{lily.h.zhang@nyu.edu}\\
      \Name{Aahlad Puli}
       \Email{aahlad@nyu.edu}\\
        \Name{Rajesh Ranganath}
       \Email{rajeshr@cims.nyu.edu}\\
       \addr Center for Data Science + Courant Institute of Mathematical Sciences \\
       New York University\\
       } 

\begin{document}

\maketitle

\begin{abstract}

In machine learning, incorporating more data is often seen as a reliable strategy for improving model performance; this work challenges that notion by demonstrating that the addition of external datasets in many cases can hurt the resulting model's performance. 
In a large-scale empirical study across combinations of four different open-source chest x-ray datasets and 9 different labels, we demonstrate that in 43\% of settings, a model trained on data from two hospitals has poorer worst group accuracy over both hospitals than a model trained on just a single hospital's data. This surprising result occurs even though the added hospital makes the training distribution more similar to the test distribution. 
We explain that this phenomenon arises from the spurious correlation that emerges between the disease and hospital, due to hospital-specific image artifacts. We highlight the trade-off one encounters when training on multiple datasets, between the obvious benefit of additional data and insidious cost of the introduced spurious correlation. In some cases, balancing the dataset can remove the spurious correlation and improve performance, but it is not always an effective strategy. We contextualize our results within the literature on spurious correlations to help explain these outcomes.
Our experiments underscore the importance of exercising caution when selecting training data for machine learning models, especially in settings where there is a risk of spurious correlations such as with medical imaging. 
The risks outlined highlight the need for careful data selection and model evaluation in future research and practice.

\end{abstract}

\section{Introduction}\label{sec:introduction}

A major challenge in machine learning, broadly and for healthcare applications, is building models that generalize under real-world clinical settings \citep{zech2018variable}. 
Collecting more labelled data from additional sources is a commonly suggested approach to improve model generalization \citep{sun2017revisiting}.
However, this work challenges that idea by highlighting where this strategy can backfire: in some situations, adding more labelled examples to the training data and training in the same manner can actually decrease model generalizability.


We demonstrate this phenomenon through a large-scale empirical study using open-source chest x-ray data from four different hospitals.
We first train individual models on data from one hospital and validate performance internally on held-out test data from the same hospital, and externally on test data from another hospital, following common practice~\citep{altman2000we,justice1999assessing}.
To measure generalization performance, we 1) group samples in the test data based on the hospital source and class label and 2) compute the worst amongst per-group accuracies, typically called worst-group accuracy.
Unlike average accuracy or AUROC, worst-group accuracy is sensitive to models trading off performance on one group for another, in turn enabling the detection of cases where one group is systematically misclassified (e.g. the diseased class from one hospital), which is especially important in healthcare applications.

\begin{figure}[t]
\vspace{-10pt}
    \centering
    \includegraphics[width=.9\linewidth]{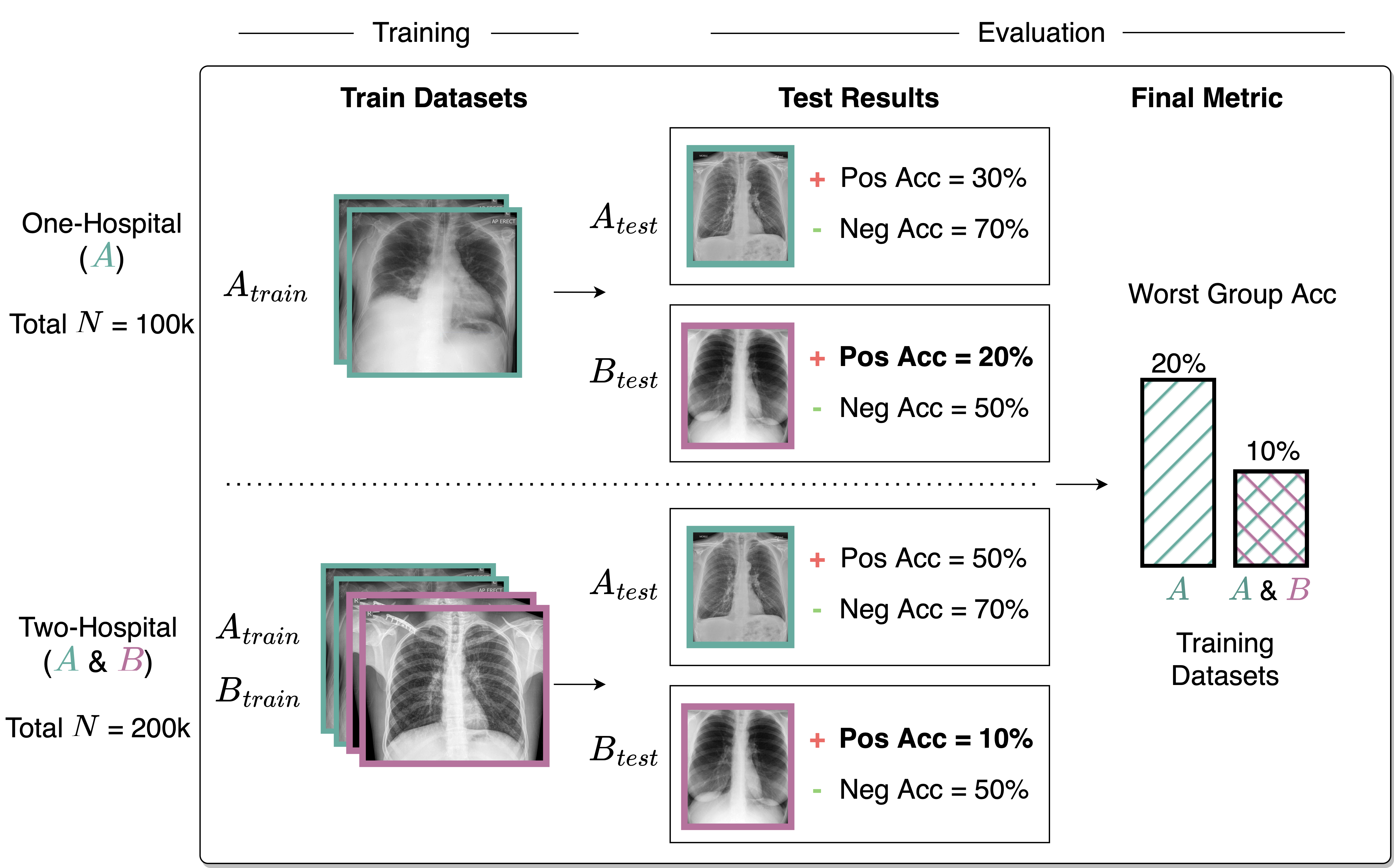}
    \caption{ High level overview of experiments. We study changes in worst-group accuracy (bold) when training on additional labelled data from an external hospital.}
    \label{fig:summary}
    \vspace{-1em}
\end{figure}

We compare the performance of models trained on data from one hospital to those trained on data from \textit{both} the original hospital and the additional external hospital.
We assess performance using the same held-out data and find that
even though the additional data \textit{comes from the exact external source we are including in evaluation}, oftentimes worst-group accuracy decreases.
See \Cref{fig:summary} for a summary of the experimental setup.


The paper is structured as follows. First, we detail the experimental setup, including our evaluation metric, worst-group accuracy (\Cref{sec:methods}). 
\Cref{sec:problem} gives the main empirical finding: in 43\% of training dataset/disease tasks, adding data from an external source hurts worst-group performance.
\emph{This result refutes the common presumption that blindly training on more
data is better for generalization \citep{sun2017revisiting}, even when the additional data brings the resulting training distribution closer to the test distribution used in evaluation.}

In the same section, we explain how this phenomenon can be understood through the lens of spurious correlations. First, adding data from an alternate source can induce a spurious correlation between the x-ray's hospital source and disease label, due to differential disease prevalences across hospitals (\Cref{sec:spurious_data}). In \Cref{sec:hospital_prediction}, we highlight why avoiding this shortcut (using hospital for disease prediction) is especially difficult, given 
how easily models pick up on hospital-specific signal in chest x-rays. In fact, we find that networks trained to predict disease often encode representations that can perfectly discriminate between hospitals that were not even seen during training. 
We also explain how models trained on such multi-source data can pick up on these shortcuts, hurting performance on groups where the shortcut pattern does not hold.

Next, we investigate a commonly proposed method for addressing spurious correlations, namely balancing disease prevalence between datasets by undersampling to remove the spurious correlation (\Cref{sec:bal}).
Balancing is often beneficial, but in the scenarios where adding an additional data source hurts generalization performance, it does not always improve generalization; in some cases, training on a balanced dataset achieves lower worst-group accuracy than training on datasets from one or two hospitals.
%
Additionally, we provide an explanation using theoretical tools \citep{puli2021nurd} to show how balancing will not always yield robust solutions. 
Our results suggest that balancing \textit{can} mitigate some of the detrimental effects of incorporating an additional data source but is not a panacea to be used blindly.

In \Cref{sec:related}, we compare our analysis and insights to existing work in spurious correlations and generalizing machine learning models on chest x-rays. We conclude with practical recommendations for approaching model building when using datasets from multiple sources.
Code to reproduce experiments, as well as full unaggregated results with other metrics for external analysis can be found \href{https://github.com/basedrhys/ood-generalization}{at this URL}\footnote{\url{https://github.com/basedrhys/ood-generalization}}.

\subsection*{Generalizable Insights about Machine Learning in the Context of Healthcare}


Our work presents the following generalizable insights for machine learning and healthcare:
\begin{enumerate}
    \item More data does not necessarily guarantee better generalization, \textit{even when the added data makes the training distribution look more like the test distribution}: in many cases, including data from other sources can hurt model generalization by introducing a new spurious correlation between data source and the label. This could occur in a realistic healthcare setting if combining data from another hospital, but also if simply from a different department (e.g., that uses different x-ray scanners). This detriment to performance happens more commonly if the added dataset has a smaller proportion of diseased instances (\Cref{fig:scatter_1e_2e})

    \item Neural networks trained to predict disease pick up on strong hospital-specific signal, enough to discriminate between hospitals that were not even part of the training data. We demonstrate this in chest x-ray disease classification and point to theory that suggests this will occur in other tasks where inputs contain hospital-specific signal.
    
    \item Balancing label prevalence between datasets, a common approach in the spurious correlations literature, does not always fix learning of spurious correlations, a result we show empirically and frame theoretically.
    
    \item Care should be taken when curating data for model building and evaluation. 
    Additional considerations include utilizing metadata in evaluations (e.g. to obtain subgroup performances), testing various shifts between train and test, and considering balancing as well as alternative algorithms that take into account its limitations for robustness \citep{puli2021nurd}. See discussion (\Cref{sec:discussion}) for more details.
\end{enumerate}


\section{Experimental setup}\label{sec:methods}

Below we detail our datasets, model and task setup, and evaluation.

\subsection{Datasets}

\begin{table}[]
\centering
\vspace{-1em}
\caption{Total number of instances and disease prevalence in each dataset.}
\begin{tabular}{lllll}
\toprule
Target Label & MIMIC & CXP & NIH & PAD \\
\midrule
Pneumonia & \cellcolor[HTML]{EEF7F3}6.82\% & \cellcolor[HTML]{F8FBFC}2.43\% & \cellcolor[HTML]{FAFCFE}1.31\% & \cellcolor[HTML]{F3F8F7}4.84\% \\
Cardiomegaly & \cellcolor[HTML]{D8EEE0}17.05\% & \cellcolor[HTML]{E2F2E9}12.38\% & \cellcolor[HTML]{F8FBFB}2.51\% & \cellcolor[HTML]{E9F5EF}9.15\% \\
Edema & \cellcolor[HTML]{E3F2EA}11.83\% & \cellcolor[HTML]{C4E6CF}26.01\% & \cellcolor[HTML]{F9FBFC}2.11\% & \cellcolor[HTML]{FAFCFE}1.23\% \\
Effusion & \cellcolor[HTML]{CBE8D4}23.18\% & \cellcolor[HTML]{A5D9B4}40.28\% & \cellcolor[HTML]{E3F2EA}11.94\% & \cellcolor[HTML]{F0F7F5}5.99\% \\
Atelectasis & \cellcolor[HTML]{D1EBDA}20.11\% & \cellcolor[HTML]{DBEFE3}15.47\% & \cellcolor[HTML]{E7F4ED}10.33\% & \cellcolor[HTML]{F1F8F6}5.50\% \\
Pneumothorax & \cellcolor[HTML]{F4F9F8}4.19\% & \cellcolor[HTML]{E9F5EF}9.25\% & \cellcolor[HTML]{F3F9F7}4.66\% & \cellcolor[HTML]{FCFCFF}0.31\% \\
Consolidation & \cellcolor[HTML]{F3F9F7}4.67\% & \cellcolor[HTML]{EEF7F3}6.81\% & \cellcolor[HTML]{F4F9F8}4.19\% & \cellcolor[HTML]{FAFBFD}1.56\% \\
Any & \cellcolor[HTML]{8ED0A0}50.73\% & \cellcolor[HTML]{63BE7B}70.35\% & \cellcolor[HTML]{C0E4CB}28.04\% & \cellcolor[HTML]{CBE8D5}23.03\% \\
No Finding & \cellcolor[HTML]{B1DEBF}34.76\% & \cellcolor[HTML]{EAF5EF}8.98\% & \cellcolor[HTML]{88CD9B}53.65\% & \cellcolor[HTML]{AEDDBC}36.12\% \\
\midrule
Num Instances & 243k & 192k & 113k & 100k \\
\bottomrule
\end{tabular}
\label{tab:dataset-stats}
\end{table}

We use four open-source disease classification datasets in this research: MIMIC-CXR-JPG (\verb|MIMIC|) \citep{johnson2019mimic}, CheXpert (\verb|CXP|) \citep{irvin2019chexpert}, Chest X-ray8 (\verb|NIH|) \citep{wang2017nih}, and PadChest (\verb|PAD|) \citep{bustos2020padchest}, filtering to include only frontal (PA/AP) images.
Instances are labeled with one or more pathologies.
Each dataset has a different set of diseases but we preprocess them using code derived from \verb|ClinicalDG|\footnote{\url{https://github.com/MLforHealth/ClinicalDG}}\citep{zhang2021empirical} to extract the eight common labels (\cref{tab:dataset-stats}) and homogenize the datasets.
Additionally, we create the \textit{Any} label which indicates a positive label for any of the seven common disease labels, resulting in \textbf{nine different binary labels}. Table \ref{tab:dataset-stats} gives an overview of the relative sizes and prevalences of each disease for each dataset after preprocessing. All experiments use the labels in a binary manner; a pathology is chosen as the target label, with an instance labeled \verb|1| if the pathology of interest is present and \verb|0| otherwise. 
We apply an 80\%/10\%/10\% subject-wise train/val/test split, with the same split used across seeds.

Each dataset is designated as its own \textit{domain}. We use the terms \textit{domain} / \textit{hospital} / \textit{environment} interchangeably, each referring to a specific dataset (\verb|MIMIC|, \verb|CXP|, \verb|NIH|, \verb|PAD|) or dataset combination (\verb|MIMIC+CXP|, \verb|MIMIC+NIH|, \verb|MIMIC+PAD|, \verb|CXP+NIH|, \verb|CXP+PAD|, \verb|NIH+PAD|). Across single- and double-dataset configurations, we have \textbf{ten total dataset configurations} used throughout this research. 

\subsection{Model \& Task Setup}\label{sub:model_task_setup}

As it is shown to be strong baseline for chest x-ray classification \citep{bressem2020comparing, raghu2019transfusion}, we use the same model architecture as \citet{zhang2021empirical}: a DenseNet-121 network \citep{huang2017densenet} initialized with pre-trained weights from ImageNet \citep{imagenet}. We replace the final layer with a two-output linear layer (for binary classification). For simplicity, we only consider binary disease classification. 
For training the network, all images are resized to 224 $\times$ 224 and normalized to the ImageNet \citep{imagenet} mean and standard deviation. During training we apply the following image augmentations: random horizontal flip, random rotation up to 10 degrees, and a crop of random size (75\% - 100\%) and aspect ratio (3/4 to 4/3). All runs use \verb|Adam| with \verb|lr| = 1e-5 and \verb|batch size = 128|, which was found to be a performant configuration in early tuning (\citep{zhang2021empirical} use \verb|lr| = 5e-4 and \verb|batch size = 32|). Training runs for a maximum of 20k steps, with validation occurring every 500 steps and an early stopping patience of 10 validations. All test results are obtained using the optimal model found during training as measured by the highest validation macro-F1 score (following \citep{fiorillo2021deepsleepnet, berenguer2022macrof1}) as it gives a robust ranking of model performance under imbalanced labels. \Cref{fig:es_steps} visualises the number of training steps chosen by early stopping for all different models, showing that performance saturates well before the 20k step limit.

\subsection{Evaluation}

Following recommendations to perform external validation alongside internal validation \citet{justice1999assessing,altman2000we}, for each base hospital we choose one additional hospital to include in evaluation, e.g., evaluating a model trained on \verb|MIMIC| data using \verb|MIMIC| and \verb|PAD| data. To provide a more fine-grained analysis of model performance, we analyse accuracies within each class for each hospital. The result is four different sub-population accuracies for each binary classification: a group for the disease class from hospital A, the non-disease class from hospital A, the disease class from hospital B, and the non-disease class from hospital B. To emphasize our focus on robustness, we report the worst accuracy of the four groups. 
Worst-group accuracy across classes in internal and external sites describes the reliability of a trained model in deployment (which models have been shown to struggle with especially in a healthcare setting \citep{subbaswamy2020deployment}); it can only be high when a model is performing well on all classes both internally and externally.
In contrast to worst-group accuracy, aggregate measures such as overall accuracy can be high even if one group has low performance, or AUROC which would not pick up on differential performance between different sites.
This focus on the performance of individual groups is especially important in healthcare given that minority groups are important even though they are small proportionally (e.g., the disease group from a given hospital where missing a diagnosis can prove fatal).


\section{The Dangers of Combining Data}
\label{sec:problem}

\begin{figure}[t]
    \centering
    \includegraphics[width=\linewidth]{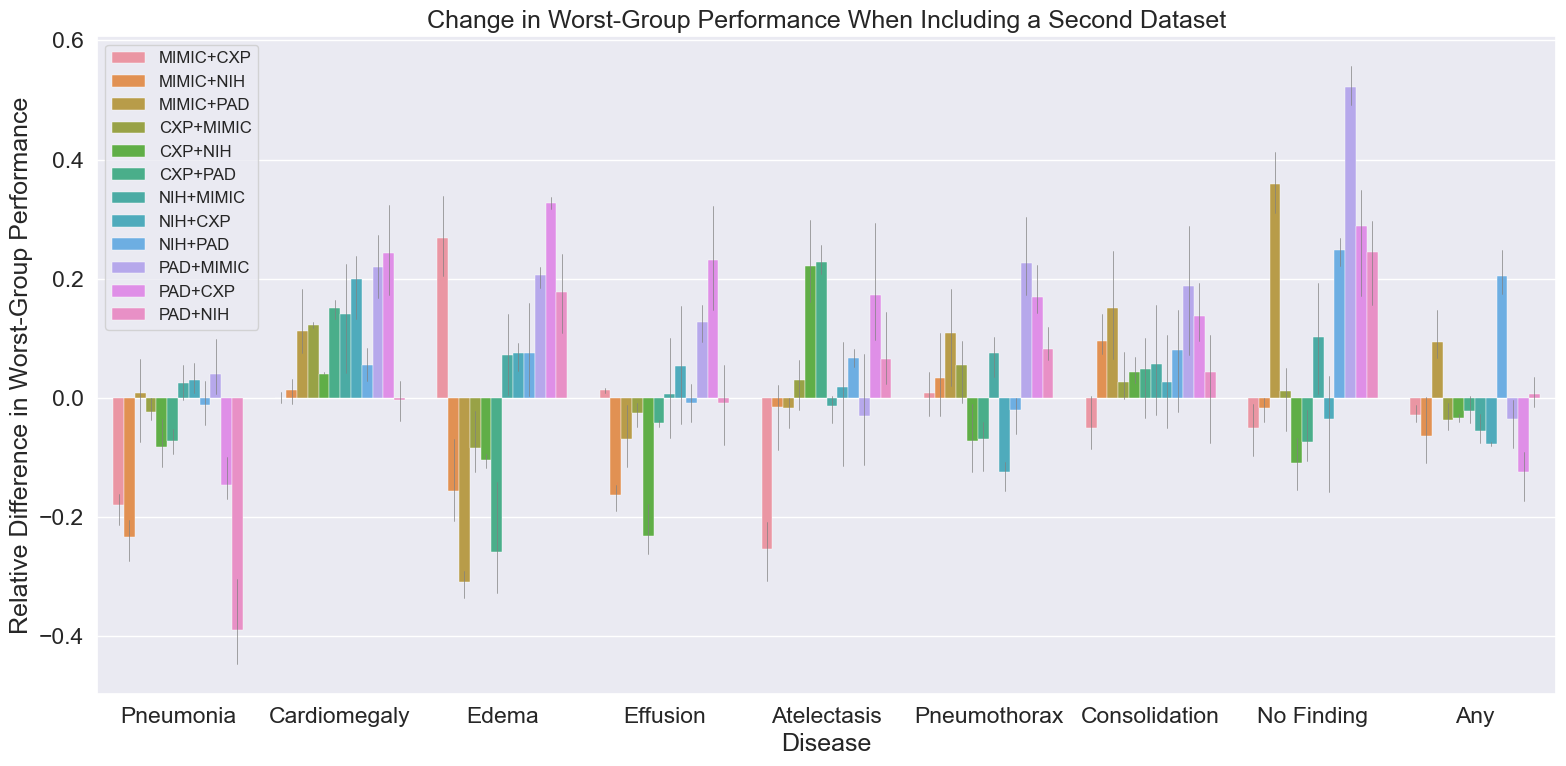}
    \caption{Change in worst-group performance after including a second training dataset. We see both improvements in performance as expected, but also many deteriorations. For every element in the legend, the first dataset is the base environment, and the second is the added environment; for instance, MIMIC+CXP shows the performance change from training on just MIMIC to training on MIMIC and CXP.}
    \label{fig:adding_second_dataset}
\end{figure}

Here, we evaluate the impact of incorporating data from an additional hospital on model performance. Namely, we train a separate model on each combination of dataset and target label possible out of all ten single- and multi-source dataset combinations and all nine different target labels, with three seeds for each configuration.
We report the change in worst-group accuracy between a model trained on the original single-source data and a model trained on the larger multi-source dataset, where the added environment matches the environment used as the external site in evaluation. Figure \ref{fig:adding_second_dataset} summarizes these results, showing the change in worst-group accuracy when an external dataset is added. For completeness, we also show absolute AUROC values in Appendix \Cref{fig:absolute_auroc} for the one-environment and two-environment models.

Our results indicate that adding external data from the same distribution as used in evaluation can actually harm worst-group performance.
In 43\% of cases, the addition of an external dataset leads to a decrease in worst-group performance.
While we do see improvements in worst-group accuracy in some cases, performance improvement is far from guaranteed. Moreover, recall that the data we are adding is from the external site used for evaluation and is thus a best-case scenario data source (i.e. relative to augmenting with some other external dataset). The fact that worst-group performance decreases so often even in this setting suggests that incorporating this additional data for training is introducing a harm that outweighs the gains one would otherwise expect when including data similar to the held-out data used in evaluation (thus making the training data more similar to test). 
\textbf{These results refute common wisdom that training on more data will help generalization performance even when the new dataset is more similar in distribution to the test data than the original training dataset.}

Given these surprising results, we aim to understand what is causing the decreases in performance after including the external dataset; we claim that the drops in worst-group performance are due to introduced spurious correlations, discussed below.

\subsection{How Adding Data Can Induce Spurious Correlations}
\label{sec:spurious_data}
Here, we explain how combining data from two sources can introduce a \textit{spurious correlation} that models exploit, resulting in lower worst-group performance. Spurious correlations are relationships in a particular data distribution that do not necessarily hold over time or reflect a genuine connection \citep{Geirhos2020ShortcutLI}. In the context of medical imaging, spurious correlations may exist between the disease of interest and non-physiological signals such as scanner-specific artifacts \citep{badgeley2019hipfracture}, simply due to hospital-specific processes that send certain patient types to certain scanners / departments. Spurious correlations are often a consequence of the specific data generating process of a given training set, and here we explain how combining data sources itself can lead to spurious correlations in the resulting training data.


%
Recall from \Cref{tab:dataset-stats} that disease prevalences can differ substantially between hospitals (e.g. the proportion of disease can be several times higher in one hospital than another). This difference in disease prevalence means that the resulting dataset after combining multiple hospitals's data exhibits a correlation between the hospital source and the probability of disease. We call this correlation a spurious correlation because it does not reflect a true physiological relationship one would wish to exploit when diagnosing disease and provides no predictive power within each hospital.

It is worth noting that there technically exists a hospital-disease correlation in all hospital pairs we test, due to the differential disease prevalences across hospitals. However, some correlations are much smaller than others, and the existence of a correlation in the training data does not guarantee that a model trained on this data will pick up on the correlation. Despite this caveat, we provide evidence that many of the models trained on combined data sources \textit{do} pick up on these hospital-disease spurious correlations; we will do so by comparing their worst-group accuracies with their single-source counterparts. 




\paragraph{Group accuracy changes align with the use of the hospital shortcut.}

Here, we aim to understand if models are indeed using this hospital-specific information in their predictions; we do this by looking at how the prediction accuracies change across groups upon the introduction of an additional dataset. To make the analysis easier, we first define the concept of a \textit{shortcut group} and \textit{leftover group}. The \textit{shortcut group} consists of instances that would be correctly labeled by the shortcut; as an example, if hospital A has a higher prevalence of disease than hospital B, then a shortcut that predicts disease if an x-ray is from hospital A and non-disease otherwise will perfectly predict the positive class of hospital A and the negative class of hospital B; these instances comprise the \textit{shortcut group}. The \textit{leftover group} are the remaining instances --- those that a prediction based on shortcut alone would label incorrectly. In the running example, the \textit{leftover group} would consist of the non-disease instances from hospital A and positive disease instances from hospital B. 

The use of a shortcut increases the accuracy of the \textit{shortcut group} and decreases the accuracy of the \textit{leftover group}, since the shortcut gets predictions right in the \textit{shortcut group} and wrong in the \textit{leftover group}; in contrast, we would not expect the correct use of physiological features to decrease the accuracy of any group. Thus, if moving from a single-source to multi-source dataset results in better \textit{shortcut group} accuracy and worse \textit{leftover group} accuracy, we have strong evidence the model is leveraging the hospital-label spurious correlation. Of the 47 configurations with worse worst-group accuracy, we see increased \textit{shortcut group} accuracy and decreased \textit{leftover group} accuracy in 37 cases (79\% of cases).
Finally, note that improvements in worst-group accuracy do not preclude shortcut learning as other factors could provide larger improvements than the drops induced by exploiting a shortcut: for example, when the base environment has few diseased samples, the model trained on this alone may simply not have enough data to learn the physiological signal effectively, so adding external data with a large disease prevalence can improve positive class accuracy in both hospitals.

In our experiments, we find that the worst group under the multi-source trained model is part of the leftover group in a majority of cases (44 out of 47 with performance degradation, 92 out of 108 overall). In other words, the group that has the worst performance consists of instances that the shortcut would have labeled incorrectly, indicating again that the model is leveraging the hospital-label correlation, and suggesting how the use of a shortcut can lead to worst-group accuracy decreases.

We also note that the addition of an external dataset increases the total dataset size, adding a confounder to our analysis. To control for this, we perform the same analysis comparing single-source to multi-source training performance, but where the multi-source dataset is uniformly undersampled to the same size as the single-source dataset (Appendix \Cref{fig:1E_2E_both} (bottom)).
With this, we see even fewer improvements / more reductions in performance from including the external dataset; 51\% of tasks (compared to 43\% previously) see a decrease in worst-group accuracy, with the remaining 49\% of tasks seeing an improvement in worst-group performance when including the external data with dataset size controlled for. This indicates that some of the positive effect of including external datasets can be attributed simply to more training instances, however, dataset size is not the only improving factor. The remaining improvements are likely due to inclusion of the external data adding variation to the training data and making the training distribution more similar to the test distribution.

\subsection{The Costs and Benefits of Additional Data Sources are Task-Dependent}

Based on the above analysis, we note at least two competing forces when adding an additional data source to the training set. On one hand, the addition of data from a new site can improve model performance, especially on data from that site. On the other, the addition of a new data source can induce a spurious correlation between hospital and disease.

We find that the effect of adding a second source varies substantially between diseases.
For instance, both \textit{Pneumonia} to \textit{Cardiomegaly} see significant differential disease prevalence between hospitals (and so a strong potential for spurious correlations to be learned), but \textit{Cardiomegaly} worst-group accuracy is almost universally improved by additional datasets, while \textit{Pneumonia} sees few improvements and many significant degradations. This could be due to the relative hospital- vs disease-signal difference exhibited by these two diseases; \textit{Cardiomegaly} is easier to predict than \textit{Pneumonia} (median worst-group accuracy of 0.36±0.1 vs 0.16±0.1),
and so a model does not need to rely on hospital-specific signal during training. In contrast, \textit{Pneumonia} is harder to predict, so a model may benefit more from relying on hospital-specific signal during training.
There could be a myriad of other reasons for this variability (e.g., inter-hospital concept shift where labels differ between datasets due to differing labelling mechanisms\citep{cohen2020limits}), and while diagnosing the exact cause is outside the scope of the paper, our results emphasize that performance improvements due to including external data are disease-dependent and so it is important to take into account the specific disease of interest when considering data to use for model-building. 

\begin{figure}[t]
    \centering
    \includegraphics[width=.7\linewidth]{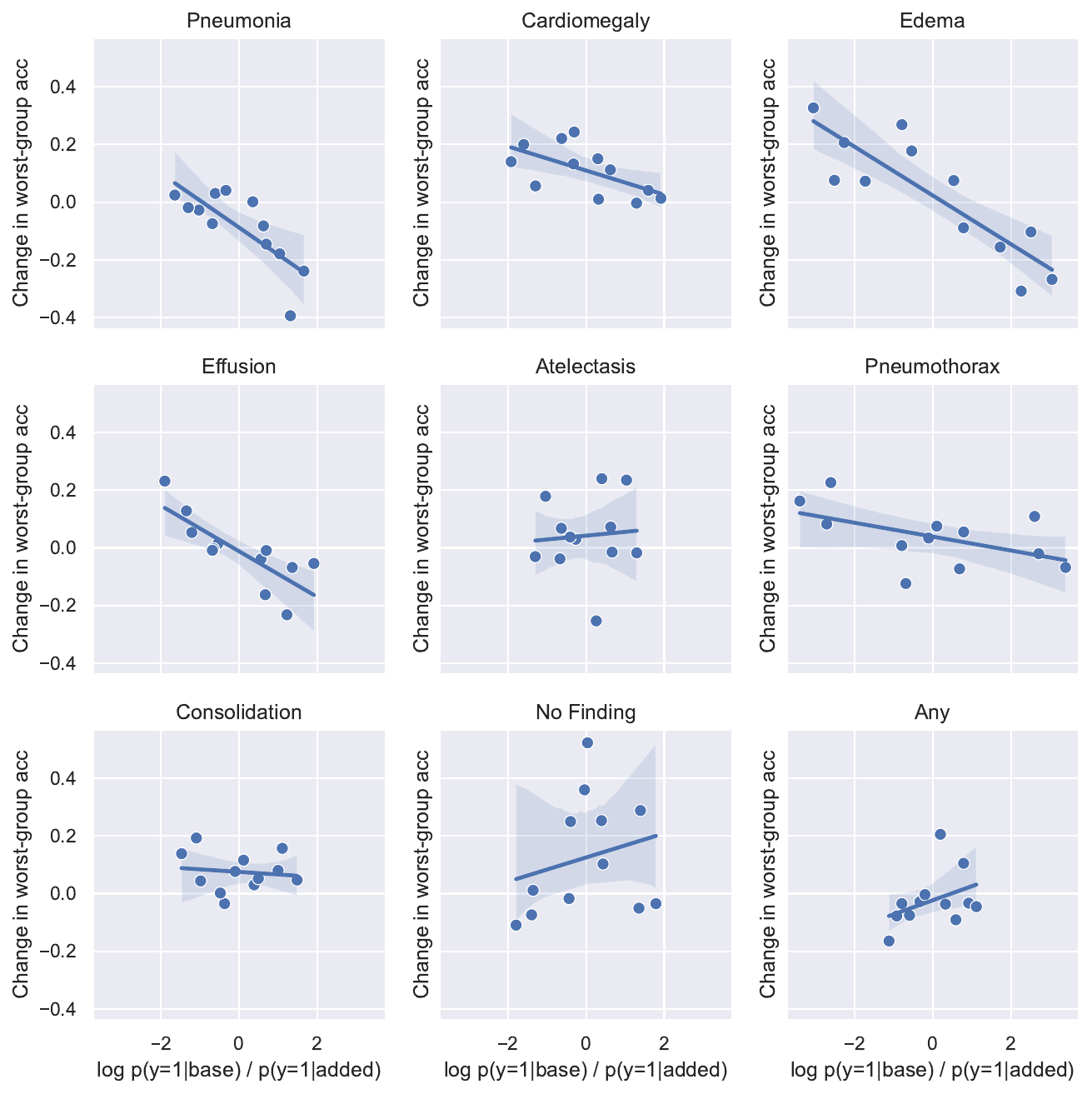}
    \caption{ For most tasks, the change in worst-group accuracy when adding an additional data source tends to correlate with the log-ratio of the disease prevalence between the original and additional dataset. For most tasks that involve individual diseases, the lower the disease prevalence of the added dataset relative to the original, the more adding the dataset hurts worst-group performance.}
    \label{fig:scatter_1e_2e}
\end{figure}

We see that for most diseases, within a disease, the lower the disease prevalence in the added dataset relative to the original dataset, the worse the delta in worst-group performance between the multi-source and single-source model. In fact, a linear regression using just the log-ratio of the disease prevalence, disease, and their interactions as features yields predictions that have a Pearson correlation of .57 with the actual results (see \Cref{fig:scatter_1e_2e} for visualization). This result suggests that one should be especially wary of combining data naively from an external source whose disease prevalence is small in comparison to the original source.

\subsection{Models pick up on hospital-specific features}\label{sec:hospital_prediction}

We now elucidate the cause for models capturing these hospital-label spurious correlations so often when trained on multi-source datasets; specifically, because hospital-specific artifacts are so deeply embedded in chest x-ray images. First, we show that disease prediction model embeddings (from the same models as discussed above) encode highly discriminative hospital information, even between hospitals not seen in training. Then we show that a CNN trained directly for hospital prediction can do so with near perfect accuracy from a \textit{very small ($8\times8$) center crop}.

\paragraph{Model representations capture hospital source.}
First, we probe the information contained in the penultimate activations of the trained models. Taking embeddings from the model trained with seed 0 for every combination of the ten hospital subsets (both single- and double-env combinations) and nine diseases on each hospital's test set (undersampled to the size of smallest test set), we train a linear SVM to predict hospital source. We test the embeddings' ability to predict any combination of hospital pairs (or between all four hospitals), regardless of which hospital(s) the original model was trained on.
For each embedding and task, to remove inter-label variation, we train two linear SVMs; one on non-disease x-ray embeddings and one on disease x-ray embeddings.

We find that every CNN model, regardless of training disease or datasets, learns embeddings that can distinguish any of the hospital sources with near-perfect accuracy, even though the embeddings were trained via one or two hospitals' data. See \Cref{fig:embedding-analysis-env} in Appendix. Our results show that CNN models encode highly discriminative hospital signal \textit{even when trained to predict disease}. In contrast, the same embeddings only reach 70-80\% accuracy on disease prediction (the task they were trained on).

We also find that when training CNNs directly for hospital prediction, this hospital-specific signal exists in regions as small as \verb|8x8| pixels. \Cref{tab:hospital_prediction} shows the results of four-way hospital prediction when DenseNet-121 models are trained on varying-sized center crops, resized then to \verb|224x224| for model input. For ease of evaluation, each hospital's data is uniformly undersampled such that all hospitals are of equal size. Four-way hospital prediction can be done perfectly with a center crop of \verb|128x128| pixels, and perfectly between MIMIC/CXP and NIH/PAD with a center crop of \verb|8x8|. These results further show that hospital-specific signal is deeply encoded in chest x-rays (enough that models can pick up on it even from a very small patch of an image), explaining why CNNs trained for disease prediction are so prone to learning hospital-label shortcuts.

\begin{table}
\vspace{-1em}
    \caption{ Accuracy on four-way and two-way hospital prediction task. A CNN can predict between MIMIC/CXP and NIH/PAD with perfect accuracy down to a $8\times8$ center crop, showing that hospital-specific signal strongly exists in chest x-rays}
    \label{tab:hospital_prediction}
    \centering
\begin{tabular}{l|llllll}
\toprule
Crop Size & 224 & 128 & 64 & 32 & 16 & 8 \\
\midrule
Two Class & 1.0 & 1.0 & 1.0 & 1.0 & 1.0 & 0.996 \\
Four Class & 0.999 & 0.994 & 0.974 & 0.904 & 0.796 & 0.730 \\
\bottomrule
\end{tabular}
\end{table}


\section{Balancing Can Help but Does Not Always Improve Performance}\label{sec:bal}

A commonly proposed method to address spurious correlations is balancing the data to remove any correlation between the shortcut feature (hospital in our case) and the label; previous work shows this to be competitive with other more sophisticated inter-domain generalization methods \citep{sagawa2020overparameterization,idrissi2022simpledatabalancing}. To investigate whether balancing disease prevalence between two datasets can help mitigate the spurious correlations between disease label and hospital, we repeat a similar experiment to \Cref{sec:problem}, but compare performance between training on a single dataset and training on two datasets \textit{after balancing disease prevalence between them}; again, we assess via worst-group performance. We balance using the following heuristic: from the two unaltered datasets (Hospital A, Hospital B), choose the one with higher disease prevalence and undersample the other dataset such that $P_{Hosp~A}(Y = 1) == P_{Hosp~B}(Y = 1)$. We therefore only undersample the majority (typically negative, non-diseased) instances, as previous work shows this to be the minimax optimal strategy \citep{chatterji2022undersampling}.

\begin{figure}[t]
\small
    \centering
    \includegraphics[width=\linewidth]{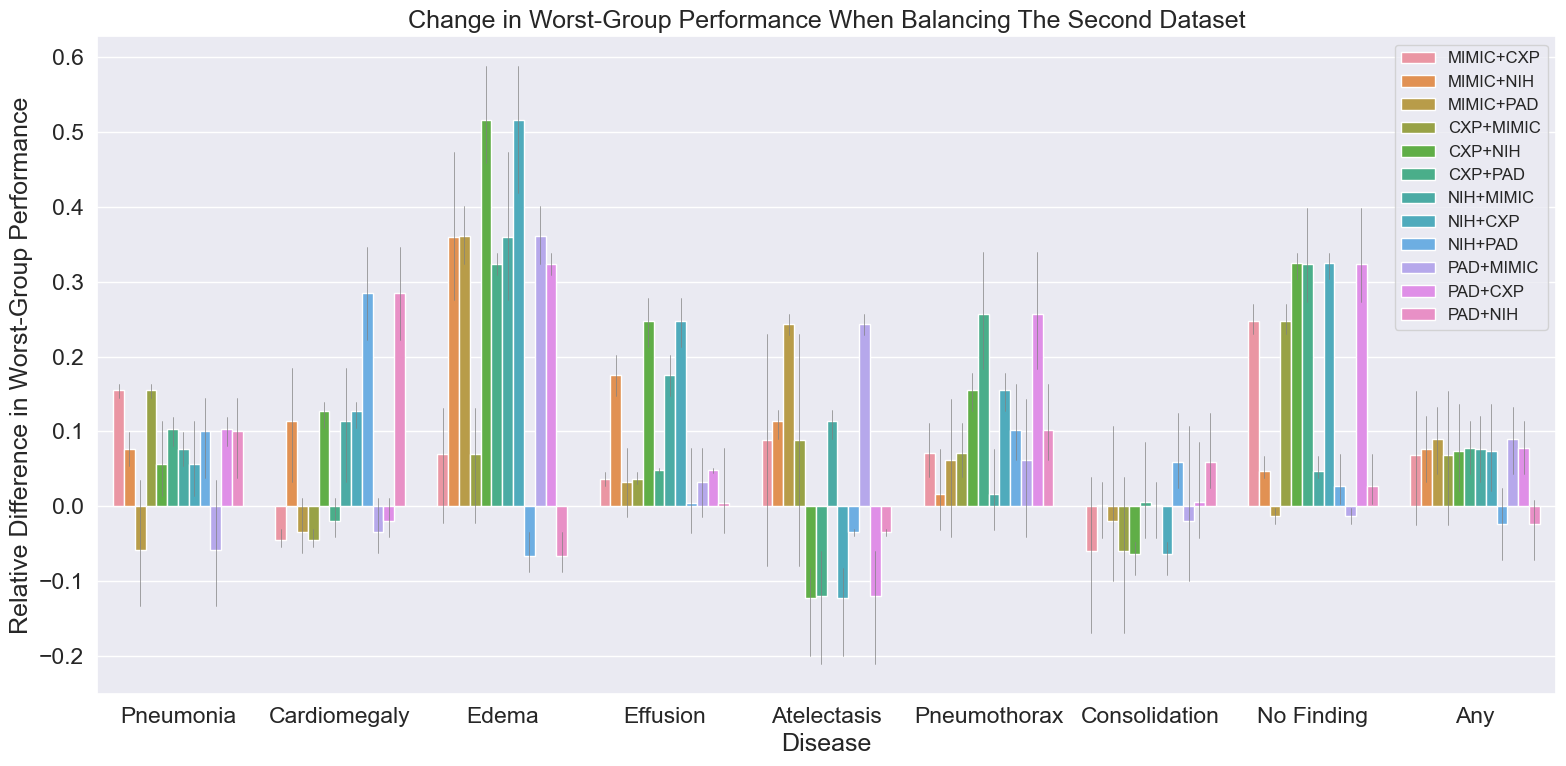}
    \caption{Change in worst-group performance when balancing the two datasets (compared to not balancing). While balancing often helps, it can also hurt performance, suggesting that it can be a promising intervention but should not be used blindly.}
    \label{fig:app_balancing}
\end{figure}

\begin{figure}
    \centering
    \vspace{-1em}
    \includegraphics[width=\linewidth]{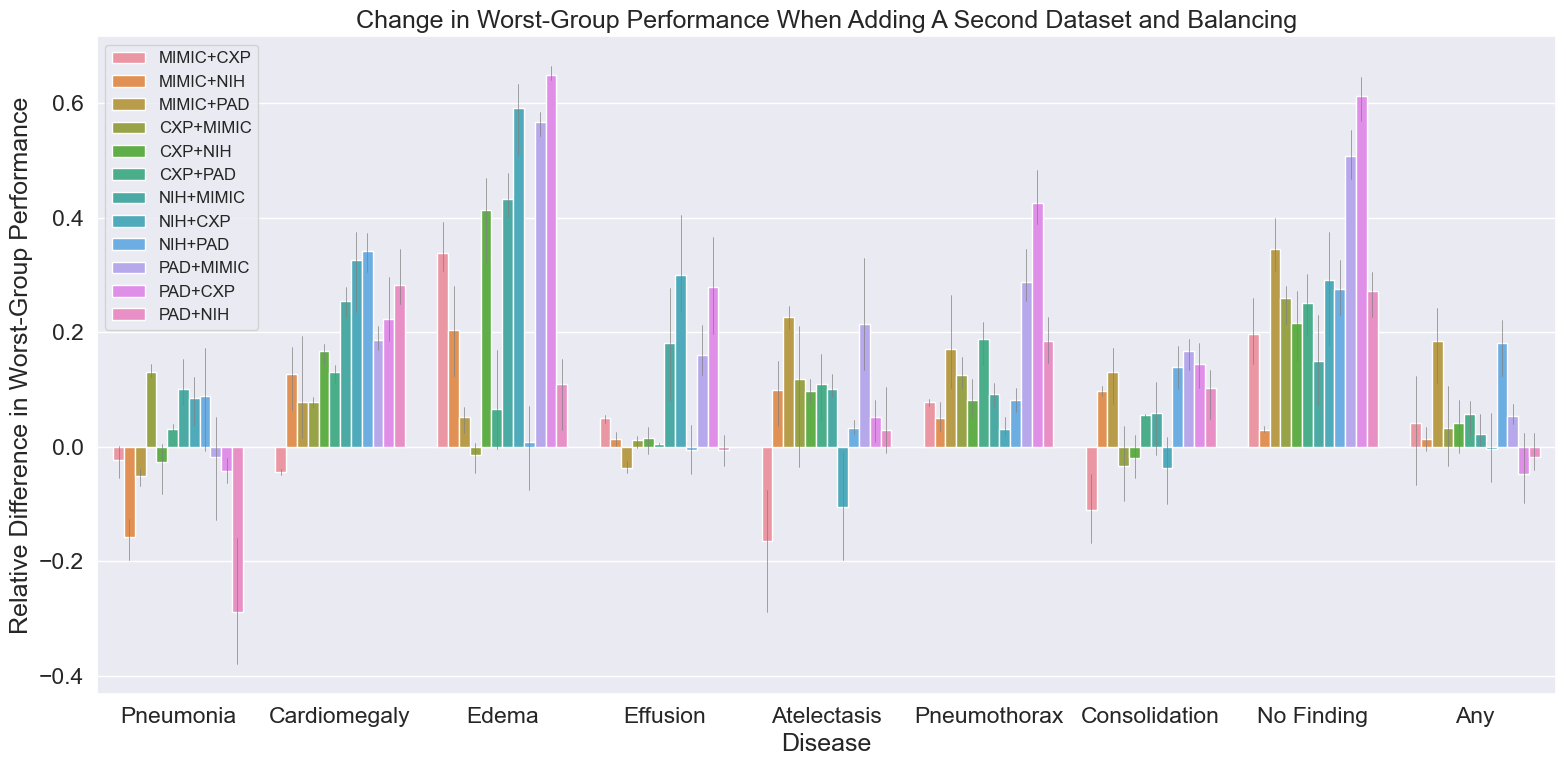}
    \caption{Change in worst-group performance after including a second dataset and balancing the two datasets. More consistently (than Figure \ref{fig:adding_second_dataset}) we see improvements but not a global increase in performance.}
    \label{fig:adding_and_balancing}
\end{figure}

Our results (\Cref{fig:app_balancing} and \Cref{fig:adding_and_balancing}) show that balancing the data to remove the direct spurious correlation in the combined dataset can often improve performance. In fact, of the 47 cases where adding additional data hurt performance, balancing the data improved performance (relative to not balancing) in 45 cases; in the other two cases, however, balancing reduced performance, suggesting that even this intervention is not \textit{guaranteed} to yield a benefit.
Despite improvements in 45 cases (relative to not balancing), only 33 of the 45 cases resulted in a two-hospital combination that outperformed using only one hospital for training. In other words, even after removing a direct correlation between hospital and disease through balancing, there are many cases where it is still better to stick with a single-source training set. 

Balancing is also not an intervention to be applied blindly, even if differential disease prevalences exist between hospitals. Of the 61 cases where incorporating additional data improved performance, balancing the combined data hurt performance relative to not balancing in 26 cases (43\%), likely due to undersampling removing data. Overall, models trained on \textit{balanced} multi-source data are better than their single-source counterparts more often than models trained on multi-source data \textit{without balancing}; however, contrary to existing works which assert that balancing is sufficient to address spurious correlations \citep{idrissi2022simpledatabalancing, kirichenko2020last}, our empirical analysis highlight that this intervention has failure modes.

One limitation of undersampling to fix spurious correlations is the loss of data; to control for this, we run the same experiment but with dataset size fixed to the size of the base hospital (Appendix \Cref{fig:2E_2EB_both} (bottom)) (as done in Section \Cref{sec:spurious_data}). As expected, when dataset size is fixed, balancing no longer results in less data than not balancing, and so is more consistently a beneficial procedure. Under this scenario, we only see $16\%$ of tasks with a decrease in worst-group accuracy, compared to the $27\%$ we saw when dataset size wasn't controlled for. These results indicate that generally, having a balanced disease proportion between hospitals is desirable, however, again, it is not \textit{always} the optimal data collection strategy. 

\begin{figure}[t]
    \centering
    \vspace{-1em}
    \includegraphics[width=.7\linewidth]{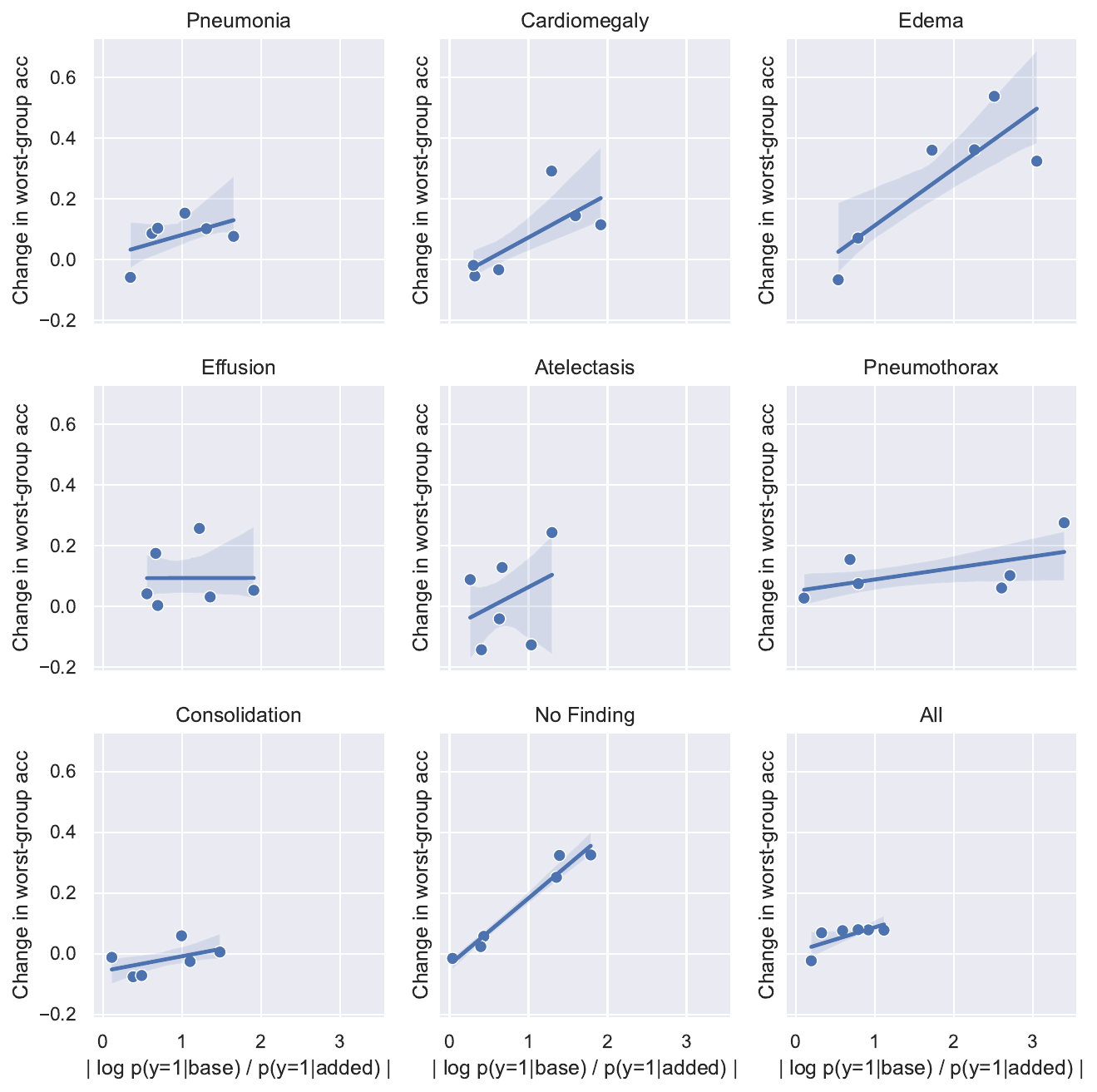}
    \caption{Within a given dataset, balancing tends to help more relative to not balancing when the difference in disease proportions between the two hospitals is greater.}
    \label{fig:scatter_2e_2eb}
\end{figure}

When is balancing most useful? We find that, within a given disease prediction task, the benefit of balancing is highest when the datasets have very different disease prevalences, regardless of which has a higher prevalence. Indeed, a linear model using just the absolute value of the log-ratio of disease prevalences, disease, and their interactions achieves predictions that have a Pearson correlation of .74 with the actual performance changes. See \Cref{fig:scatter_2e_2eb} for a visualization of the disease-dependent trends between performance changes after balancing and the ratio of the disease prevalences.

To shed further light into where balancing works, we turn to prior theoretical results \citep{puli2021nurd}. They show that balancing can fail when the shortcut feature is not perfectly predictable from the covariates, with the key insight that while balancing breaks the marginal relationship between the label and the shortcut, the label and shortcut can still be dependent after conditioning on the input, meaning models may still face performance degradation due to their dependence on hospital-specific features.
%
%
%
As an example from the experiments in \Cref{fig:app_balancing} that empirically demonstrates this, balancing fails to improve worst-group accuracy most often for the Atelactasis label and indeed, the final learned representations of the Atelactasis-prediction model are least predictive of hospital ($93\%$ vs. $99\%$).

\section{Related Work}\label{sec:related}



Our work builds on existing work in the literature on machine learning for chest x-rays and spurious correlations more broadly, which we describe below.

\paragraph{Deep Learning on Chest X-ray Data}
Our work falls within a broader line of work that considers the performance of chest x-ray models trained on data from multiple different hospital systems. \cite{zech2018variable} show that training on data from two sources improves average accuracy and AUC on a test set containing data from both sources, relative to training on either individual source alone. In contrast, we focus on worst-group accuracy (the four groups created from positive/negative instances from Hospital A/B) to draw attention to minority / more difficult groups and find that the model trained on combined data can actually exhibit worse worst-group performance. In addition, they show results for just a single pair of datasets while we consider 6 different pairs.
\cite{pooch2019domainshiftcxr} study the same chest x-ray datasets as our work, but they only consider models trained on single-source datasets.
Finally, \citet{zhang2021empirical} compare domain generalization algorithms across hospital shifts but also only consider the single-source setting. They also compare domain generalization under some spurious correlations, but their correlations are synthetically generated via random label flips and added input Gaussian noise. In contrast, we consider spurious correlations that naturally arise when combining datasets from different sources. 

Our embedding and patch analyses (\Cref{sec:hospital_prediction}) are related to existing works that highlight the ease at which hospital-specific signals can be detected. While \citep{zech2018variable} first show that convolutional neural networks can be trained to discriminate between hospital source with high accuracy, our embedding analysis goes a step further to show that the representations learned when predicting \textit{disease} are even able to discriminate between hospitals. \cite{badgeley2019hipfracture} show that the embeddings of even a randomly initialized network tend to cluster by scanner type, while we show that the embeddings of a model trained on a single hospital still contain information to discriminate between all other hospitals.



\paragraph{Spurious correlations}

Our work is also closely related to existing work on spurious correlations \citep{Geirhos2020ShortcutLI}. A unique aspect of the spurious correlation we analyze is the fact that it is induced by combining datasets 
together, in a way that matches the test setup, thus leading to our paper's main message that it is important to take care when curating data from multiple sources. \cite{idrissi2022simpledatabalancing} also question the ``just collect more data'' wisdom, though their results focus on the difference between subsampling which removes data and other algorithms for addressing spurious correlations which do not.
While they show that data balancing achieves state-of-the-art worst-group accuracies on standard spurious correlation benchmarks, they only consider balancing relative to other algorithms on full ``multi-source'' datasets. In contrast, we show that balancing can sometimes hurt performance relative to simply using the original source or training on both sources.

\paragraph{Domain Adaptation}

Our work has similarities to the domain adaptation field, which aims to build methods that generalize between domains (hospitals, in our case) \citep{csurka2017domain, motiian2017few, motiian2017unified}. One key difference between this and our work is that we look at the effect of adding data from a test domain, to make the training domain more closely match the test domain, as opposed to domain adaptation that typically looks at training on one domain while testing on another. We also evaluate across both domains simultaneously, which is important in a healthcare setting as models may be deployed in both internal and external settings and we want them to perform well under all settings.

\section{Discussion}\label{sec:discussion}


With the open-source release of multiple chest x-ray datasets, many works have endeavored to use these datasets to build models that generalize. More broadly, there is increasing attention on improving deep learning models' generalization performance by incorporating additional datasets during training. 
Given the prevalent belief in the deep learning community that gathering more data from varying sources leads to better models,
our work offers an important cautionary tale by highlighting the dangers of naively combining datasets. 
In particular, we highlight the need to consider the potential for spurious correlations when combining datasets, as well as the importance of careful evaluation beyond aggregate measures of performance. Previous work (and common wisdom) suggests that multi-source training datasets always result in more generalizable models \citep{zech2018variable}, but we show that this is far from a guaranteed outcome.

Regarding the applicability of these findings to other medical imaging domains, our theory suggests that if hospital-related signals (like hospital/scanner) can be accurately predicted (as we show in \Cref{sec:hospital_prediction}), models will be susceptible to the same degradation we present. Other findings show that hospital variables can be deduced from hip x-rays, indicating a broader application to other x-ray types~\citep{badgeley2019hipfracture}. These variables, embedded in chest x-rays due to factors like scanner type and hospital-related artifacts, are not specific to x-ray imaging only and so would likely be found in other modalities like CT and ultrasound.

We show that we can often remove the spurious correlation and address performance degradation with methods like balancing \citep{idrissi2022simpledatabalancing}, however, our results show that balancing will not always improve performance over not balancing. Moreover, balancing, even when it helps over not balancing, is not guaranteed to yield performance better than not including the additional dataset at all. These results suggest that it could be worth considering algorithms for robustness that take into account the limitations of balancing \citep{puli2021nurd}.

\paragraph{Limitations}

While our rigorous experiments illuminate the dangers of merging datasets and the constraints of balancing as a remedy for learning spurious correlations, we also acknowledge certain limitations of our work.

First, we focus only on the binary classification setting, which is a simplified representation of the multi-label setting also encountered in chest x-ray classification. While this binary setting allows us to more easily balance label proportions, looking at performance changes across all labels simultaneously (the multi-label model is optimising all disease classes during training) could provide different insights than the binary models we study.

Second, there are additional strategies we could have taken to try to further optimize the models trained on multi-source data; doing so might better approximate the choices a practitioner would make to take full advantage of additional data. For instance,
future work could employ higher capacity models (e.g., \textit{DenseNet201}) which may better utilize the increased dataset size. In addition, given source metadata during training, we could have validated early stopping using worst-group accuracy instead of the non-group aware macro-F1 score. In our experiments, we chose to keep all experimental factors in the multi-source setup the same as the single-source setup to allow for a controlled analysis of the effect of increased data alone, but it may be possible for the multi-source results to improve over what we reported if other strategies are taken alongside additional data in training.

Third, we only examine the use of under-sampling as a method to balance hospital datasets. Re-weighting, an alternative approach that can balance datasets without data loss \citep{sagawa2020overparameterization,puli2021nurd}, is not considered in our analysis; it is possible that re-weighting may lead to different outcomes when dealing with differential disease prevalence between hospitals.
However, reweighting empirically gives similar but not necessarily better results to undersampling given enough hyperparameter tuning \citep{sagawa2020overparameterization}.

Our study also does not provide a \textit{complete} characterization of when data should be combined and when balancing should be used, e.g. as a property of the disease being predicted, the disease prevalences across the data sources, etc. While our work provides important cautionary advice for practitioners, future work that more completely characterizes when to expect performance improvements and degradation could be especially helpful in guiding practitioners.

Finally, our study focuses on the chest x-ray modality only. While our findings characterize the danger of combining datasets in general and we give evidence to believe these findings would hold elsewhere \citep{badgeley2019hipfracture}, it would be worth testing how closely these results persist across other modalities and when combining more than two datasets. By doing so, researchers can gain a deeper understanding of the generalizability and applicability of our results across a broader range of tasks.



\acks{This work was funded by NIH/NHLBI Award R01HL148248, NSF Award 1922658 NRT-HDR: FUTURE Foundations, Translation, and Responsibility for Data Science, NSF CAREER Award 2145542, Optum, and the Office of Naval Research. Aahlad Puli is supported by the Apple Scholars in AI/ML PhD fellowship.}

\bibliography{main}

\begin{thebibliography}{28}
\providecommand{\natexlab}[1]{#1}
\providecommand{\url}[1]{\texttt{#1}}
\expandafter\ifx\csname urlstyle\endcsname\relax
  \providecommand{\doi}[1]{doi: #1}\else
  \providecommand{\doi}{doi: \begingroup \urlstyle{rm}\Url}\fi

\bibitem[Altman and Royston(2000)]{altman2000we}
Douglas~G Altman and Patrick Royston.
\newblock What do we mean by validating a prognostic model?
\newblock \emph{Statistics in medicine}, 19\penalty0 (4):\penalty0 453--473,
  2000.

\bibitem[Badgeley et~al.(2019)Badgeley, Zech, Oakden-Rayner, Glicksberg, Liu,
  Gale, McConnell, Percha, Snyder, and Dudley]{badgeley2019hipfracture}
Marcus~A Badgeley, John~R Zech, Luke Oakden-Rayner, Benjamin~S Glicksberg,
  Manway Liu, William Gale, Michael~V McConnell, Bethany Percha, Thomas~M
  Snyder, and Joel~T Dudley.
\newblock Deep learning predicts hip fracture using confounding patient and
  healthcare variables.
\newblock \emph{NPJ digital medicine}, 2\penalty0 (1):\penalty0 1--10, 2019.

\bibitem[Berenguer et~al.(2022)Berenguer, Mukherjee, Bossa, Deligiannis, and
  Sahli]{berenguer2022macrof1}
Abel~Diaz Berenguer, Tanmoy Mukherjee, Matias Bossa, Nikos Deligiannis, and
  Hichem Sahli.
\newblock Representation learning with information theory for covid-19
  detection.
\newblock \emph{arXiv preprint arXiv:2207.01437}, 2022.

\bibitem[Bressem et~al.(2020)Bressem, Adams, Erxleben, Hamm, Niehues, and
  Vahldiek]{bressem2020comparing}
Keno~K Bressem, Lisa~C Adams, Christoph Erxleben, Bernd Hamm, Stefan~M Niehues,
  and Janis~L Vahldiek.
\newblock Comparing different deep learning architectures for classification of
  chest radiographs.
\newblock \emph{Scientific reports}, 10\penalty0 (1):\penalty0 1--16, 2020.

\bibitem[Bustos et~al.(2020)Bustos, Pertusa, Salinas, and de~la
  Iglesia-Vay{\'a}]{bustos2020padchest}
Aurelia Bustos, Antonio Pertusa, Jose-Maria Salinas, and Maria de~la
  Iglesia-Vay{\'a}.
\newblock Padchest: A large chest x-ray image dataset with multi-label
  annotated reports.
\newblock \emph{Medical image analysis}, 66:\penalty0 101797, 2020.

\bibitem[Chatterji et~al.(2022)Chatterji, Haque, and
  Hashimoto]{chatterji2022undersampling}
Niladri~S Chatterji, Saminul Haque, and Tatsunori Hashimoto.
\newblock Undersampling is a minimax optimal robustness intervention in
  nonparametric classification.
\newblock \emph{arXiv preprint arXiv:2205.13094}, 2022.

\bibitem[Cohen et~al.(2020)Cohen, Hashir, Brooks, and
  Bertrand]{cohen2020limits}
Joseph~Paul Cohen, Mohammad Hashir, Rupert Brooks, and Hadrien Bertrand.
\newblock On the limits of cross-domain generalization in automated x-ray
  prediction.
\newblock In \emph{Medical Imaging with Deep Learning}, pages 136--155. PMLR,
  2020.

\bibitem[Csurka(2017)]{csurka2017domain}
Gabriela Csurka.
\newblock Domain adaptation for visual applications: A comprehensive survey.
\newblock \emph{arXiv preprint arXiv:1702.05374}, 2017.

\bibitem[Deng et~al.(2009)Deng, Dong, Socher, Li, Li, and Fei-Fei]{imagenet}
Jia Deng, Wei Dong, Richard Socher, Li-Jia Li, Kai Li, and Li~Fei-Fei.
\newblock Imagenet: A large-scale hierarchical image database.
\newblock In \emph{2009 IEEE Conference on Computer Vision and Pattern
  Recognition}, pages 248--255, 2009.
\newblock \doi{10.1109/CVPR.2009.5206848}.

\bibitem[Fiorillo et~al.(2021)Fiorillo, Favaro, and
  Faraci]{fiorillo2021deepsleepnet}
Luigi Fiorillo, Paolo Favaro, and Francesca~Dalia Faraci.
\newblock Deepsleepnet-lite: A simplified automatic sleep stage scoring model
  with uncertainty estimates.
\newblock \emph{IEEE transactions on neural systems and rehabilitation
  engineering}, 29:\penalty0 2076--2085, 2021.

\bibitem[Geirhos et~al.(2020)Geirhos, Jacobsen, Michaelis, Zemel, Brendel,
  Bethge, and Wichmann]{Geirhos2020ShortcutLI}
Robert Geirhos, J{\"o}rn-Henrik Jacobsen, Claudio Michaelis, Richard~S. Zemel,
  Wieland Brendel, Matthias Bethge, and Felix Wichmann.
\newblock Shortcut learning in deep neural networks.
\newblock \emph{Nature Machine Intelligence}, 2:\penalty0 665 -- 673, 2020.

\bibitem[Huang et~al.(2017)Huang, Liu, Van Der~Maaten, and
  Weinberger]{huang2017densenet}
Gao Huang, Zhuang Liu, Laurens Van Der~Maaten, and Kilian~Q Weinberger.
\newblock Densely connected convolutional networks.
\newblock In \emph{Proceedings of the IEEE conference on computer vision and
  pattern recognition}, pages 4700--4708, 2017.

\bibitem[Idrissi et~al.(2022)Idrissi, Arjovsky, Pezeshki, and
  Lopez-Paz]{idrissi2022simpledatabalancing}
Badr~Youbi Idrissi, Martin Arjovsky, Mohammad Pezeshki, and David Lopez-Paz.
\newblock Simple data balancing achieves competitive worst-group-accuracy.
\newblock In \emph{Conference on Causal Learning and Reasoning}, pages
  336--351. PMLR, 2022.

\bibitem[Irvin et~al.(2019)Irvin, Rajpurkar, Ko, Yu, Ciurea-Ilcus, Chute,
  Marklund, Haghgoo, Ball, Shpanskaya, et~al.]{irvin2019chexpert}
Jeremy Irvin, Pranav Rajpurkar, Michael Ko, Yifan Yu, Silviana Ciurea-Ilcus,
  Chris Chute, Henrik Marklund, Behzad Haghgoo, Robyn Ball, Katie Shpanskaya,
  et~al.
\newblock Chexpert: A large chest radiograph dataset with uncertainty labels
  and expert comparison.
\newblock In \emph{Proceedings of the AAAI conference on artificial
  intelligence}, volume~33, pages 590--597, 2019.

\bibitem[Johnson et~al.(2019)Johnson, Pollard, Greenbaum, Lungren, Deng, Peng,
  Lu, Mark, Berkowitz, and Horng]{johnson2019mimic}
Alistair~EW Johnson, Tom~J Pollard, Nathaniel~R Greenbaum, Matthew~P Lungren,
  Chih-ying Deng, Yifan Peng, Zhiyong Lu, Roger~G Mark, Seth~J Berkowitz, and
  Steven Horng.
\newblock Mimic-cxr-jpg, a large publicly available database of labeled chest
  radiographs.
\newblock \emph{arXiv preprint arXiv:1901.07042}, 2019.

\bibitem[Justice et~al.(1999)Justice, Covinsky, and
  Berlin]{justice1999assessing}
Amy~C Justice, Kenneth~E Covinsky, and Jesse~A Berlin.
\newblock Assessing the generalizability of prognostic information.
\newblock \emph{Annals of internal medicine}, 130\penalty0 (6):\penalty0
  515--524, 1999.

\bibitem[Kirichenko et~al.(2023)Kirichenko, Izmailov, and
  Wilson]{kirichenko2020last}
Polina Kirichenko, Pavel Izmailov, and Andrew~Gordon Wilson.
\newblock Last layer re-training is sufficient for robustness to spurious
  correlations.
\newblock \emph{ICLR}, 2023.

\bibitem[Motiian et~al.(2017{\natexlab{a}})Motiian, Jones, Iranmanesh, and
  Doretto]{motiian2017few}
Saeid Motiian, Quinn Jones, Seyed Iranmanesh, and Gianfranco Doretto.
\newblock Few-shot adversarial domain adaptation.
\newblock \emph{Advances in neural information processing systems}, 30,
  2017{\natexlab{a}}.

\bibitem[Motiian et~al.(2017{\natexlab{b}})Motiian, Piccirilli, Adjeroh, and
  Doretto]{motiian2017unified}
Saeid Motiian, Marco Piccirilli, Donald~A Adjeroh, and Gianfranco Doretto.
\newblock Unified deep supervised domain adaptation and generalization.
\newblock In \emph{Proceedings of the IEEE international conference on computer
  vision}, pages 5715--5725, 2017{\natexlab{b}}.

\bibitem[Pooch et~al.(2019)Pooch, Ballester, and
  Barros]{pooch2019domainshiftcxr}
Eduardo~HP Pooch, Pedro~L Ballester, and Rodrigo~C Barros.
\newblock Can we trust deep learning models diagnosis? the impact of domain
  shift in chest radiograph classification.
\newblock \emph{arXiv preprint arXiv:1909.01940}, 2019.

\bibitem[Puli et~al.(2022)Puli, Zhang, Oermann, and Ranganath]{puli2021nurd}
Aahlad~Manas Puli, Lily~H Zhang, Eric~Karl Oermann, and Rajesh Ranganath.
\newblock Out-of-distribution generalization in the presence of
  nuisance-induced spurious correlations.
\newblock In \emph{International Conference on Learning Representations}, 2022.

\bibitem[Raghu et~al.(2019)Raghu, Zhang, Kleinberg, and
  Bengio]{raghu2019transfusion}
Maithra Raghu, Chiyuan Zhang, Jon Kleinberg, and Samy Bengio.
\newblock Transfusion: Understanding transfer learning for medical imaging.
\newblock \emph{Advances in neural information processing systems}, 32, 2019.

\bibitem[Sagawa et~al.(2020)Sagawa, Raghunathan, Koh, and
  Liang]{sagawa2020overparameterization}
Shiori Sagawa, Aditi Raghunathan, Pang~Wei Koh, and Percy Liang.
\newblock An investigation of why overparameterization exacerbates spurious
  correlations.
\newblock In \emph{International Conference on Machine Learning}, pages
  8346--8356. PMLR, 2020.

\bibitem[Subbaswamy and Saria(2020)]{subbaswamy2020deployment}
Adarsh Subbaswamy and Suchi Saria.
\newblock From development to deployment: dataset shift, causality, and
  shift-stable models in health ai.
\newblock \emph{Biostatistics}, 21\penalty0 (2):\penalty0 345--352, 2020.

\bibitem[Sun et~al.(2017)Sun, Shrivastava, Singh, and Gupta]{sun2017revisiting}
Chen Sun, Abhinav Shrivastava, Saurabh Singh, and Abhinav Gupta.
\newblock Revisiting unreasonable effectiveness of data in deep learning era.
\newblock In \emph{Proceedings of the IEEE international conference on computer
  vision}, pages 843--852, 2017.

\bibitem[Wang et~al.(2017)Wang, Peng, Lu, Lu, Bagheri, and
  Summers]{wang2017nih}
Xiaosong Wang, Yifan Peng, Le~Lu, Zhiyong Lu, Mohammadhadi Bagheri, and
  Ronald~M Summers.
\newblock Chestx-ray8: Hospital-scale chest x-ray database and benchmarks on
  weakly-supervised classification and localization of common thorax diseases.
\newblock In \emph{Proceedings of the IEEE conference on computer vision and
  pattern recognition}, pages 2097--2106, 2017.

\bibitem[Zech et~al.(2018)Zech, Badgeley, Liu, Costa, Titano, and
  Oermann]{zech2018variable}
John~R Zech, Marcus~A Badgeley, Manway Liu, Anthony~B Costa, Joseph~J Titano,
  and Eric~Karl Oermann.
\newblock Variable generalization performance of a deep learning model to
  detect pneumonia in chest radiographs: a cross-sectional study.
\newblock \emph{PLoS medicine}, 15\penalty0 (11):\penalty0 e1002683, 2018.

\bibitem[Zhang et~al.(2021)Zhang, Dullerud, Seyyed-Kalantari, Morris, Joshi,
  and Ghassemi]{zhang2021empirical}
Haoran Zhang, Natalie Dullerud, Laleh Seyyed-Kalantari, Quaid Morris, Shalmali
  Joshi, and Marzyeh Ghassemi.
\newblock An empirical framework for domain generalization in clinical
  settings.
\newblock In \emph{Proceedings of the Conference on Health, Inference, and
  Learning}, pages 279--290, 2021.

\end{thebibliography}

\clearpage

\appendix

\section{Extra Experiment Details}

\subsection{Embedding Analysis}

\begin{figure}[h]
    \centering
    \includegraphics[width=0.8\linewidth]{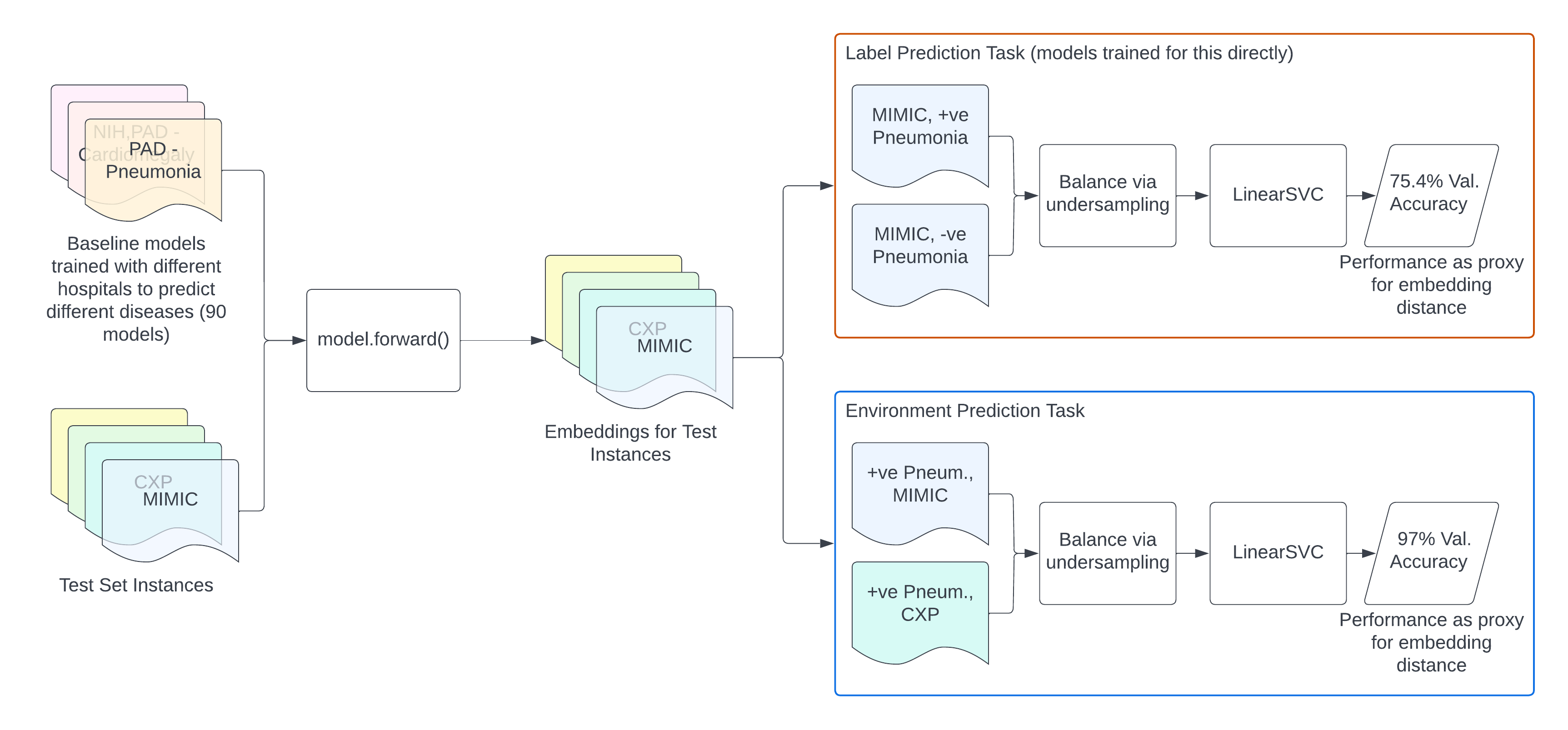}
    \caption{Overview of the Embedding Analysis process}
    \label{fig:embedding-analysis}
\end{figure}

\begin{table}[h]
\makebox[\textwidth][c]{
\begin{tabular}{l|ll}
\toprule
Task & Fix Values & Pred Values \\ \midrule
Disease Prediction & CXP, MIMIC, NIH, PAD & (0,1) \\
Environment Prediction & 0, 1 & \begin{tabular}[c]{@{}l@{}}(CXP,NIH), (CXP,PAD), (MIMIC,CXP), \\ (MIMIC,NIH), (MIMIC,PAD), (NIH,PAD), \\ (CXP, MIMIC, NIH, PAD)\end{tabular} \\ \bottomrule
\end{tabular}
}
\caption{Task configurations for embedding analysis}
\label{tab:embedding-analysis-configs}
\end{table}

Figure \ref{fig:embedding-analysis} shows the high-level process for our embedding analysis. We take all 90 baseline models with seed \verb|0| and use each to create embeddings for the test instances in each of the four environments (\verb|MIMIC|, \verb|CXP|, \verb|NIH|, \verb|PAD|). Test instances are labelled with the binary label the model was trained on and the environment the instance came from. For each of the 90 sets of test embeddings, we perform two separate tasks: classify either hospital or disease (keeping the other constant), using a linear SVM. We define two attributes: the \textbf{fixed} and \textbf{predicted} value (\textit{fix value}/\textit{pred values}). The \textit{fix value} represents the attribute that is \textit{fixed} for the SVM's dataset, while the \textit{pred value} represents the 2-4 values that the SVM is trying to predict between. The role of these two attributes are best explained with some examples.

Let's use a \textit{Pneumonia} prediction model trained on \verb|CXP|. In \textbf{environment prediction}, we might set the \textit{fix value} to \verb|Pneumonia=1| and \textit{pred values} to \verb|(MIMIC,NIH)| --- we filter the test instances to only be labelled with \textit{Pneumonia} and the SVM is classifying instance embeddings between \verb|MIMIC| and \verb|NIH|. The need for a \textit{fix value} is now clear; if we included all (both positive and negative) samples from \verb|MIMIC| and \verb|NIH|, it would be difficult to interpret the resulting SVM performance and disentangle the change in embeddings due to hospital vs change due to disease. 

For \textbf{disease prediction}, we might set the \textit{fix value} to \verb|MIMIC| and \textit{pred values} to \verb|Pneumonia=(0,1)| --- we filter the test instances to only come from \verb|MIMIC| and the SVM is classifying instance embeddings between \verb|Pneumonia=0| and \verb|Pneumonia=1|. The specific data configurations used are outlined in Table \ref{tab:embedding-analysis-configs}.

The instances are randomly subsampled between classes (e.g., between \verb|MIMIC| and \verb|CXP| for environment prediction) such that $P(Y=k) = 1 / K$ where $K$ is the number of classes. Because some hospital/disease combinations have very limited  positive instances, the resulting balanced dataset is very small -- we do not show balanced datasets of $N < ...$. After the steps outlined above, we classify the \textit{pred values} using a Linear SVM. The performance of this SVM gives a notion of the separability of the embeddings and can be used as a means to understand the semantics encoded within.

We show very high hospital prediction performance based on embeddings in \Cref{sec:hospital_prediction}, but we want to provide additional sanity checks to improve the validity of these results.

We note that performing environment prediction for a single hospital (e.g., predict which hospital an instance came from, using only \verb|CXP|) gave \textasciitilde50\% validation accuracy. This is not a noteworthy result in itself but simply a sanity check to ensure that the performance achieved by the SVM did in-fact represent some semantics encoded by the disease classification model and not simply due to data leakage/erroneous evaluation/doing prediction on such high-dimension vectors (embeddings are of size 1024).

\clearpage

\section{Extra Results \& Figures}

\subsection{Training Steps per Model}

\begin{figure}[h]
    \centering
    \includegraphics[width=0.8\linewidth]{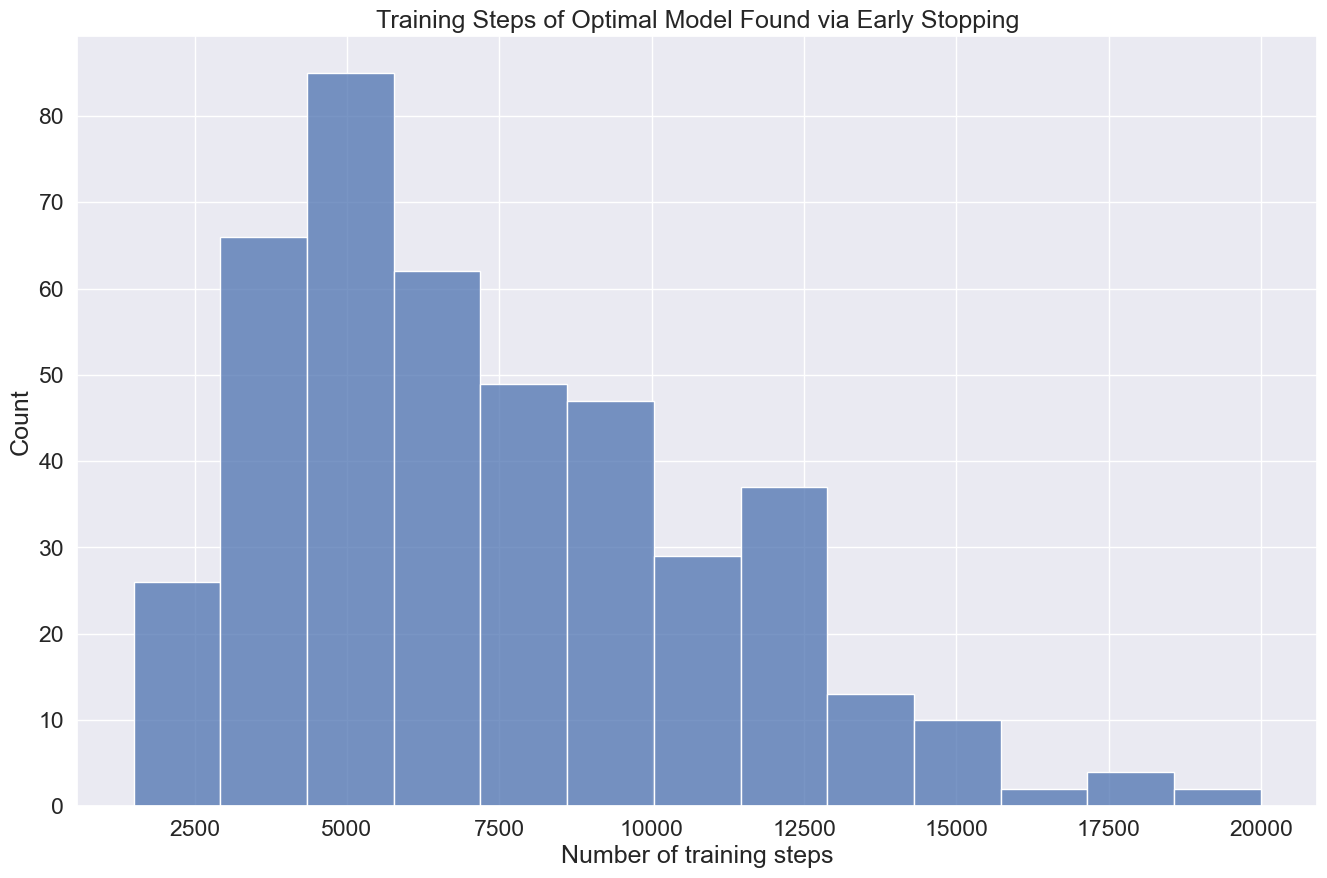}
    \caption{Histogram of number of training steps for different models trained, as chosen by early stopping. Performance typically saturates very early, as shown by the left skew of the distribution. This shows that training for longer is unlikely to give different results (at least with the data augmentation methods and DenseNet-121 model used in our work)}
    \label{fig:es_steps}
\end{figure}

\subsection{Absolute AUROC}

\begin{figure}
    \centering
    \includegraphics[width=\linewidth]{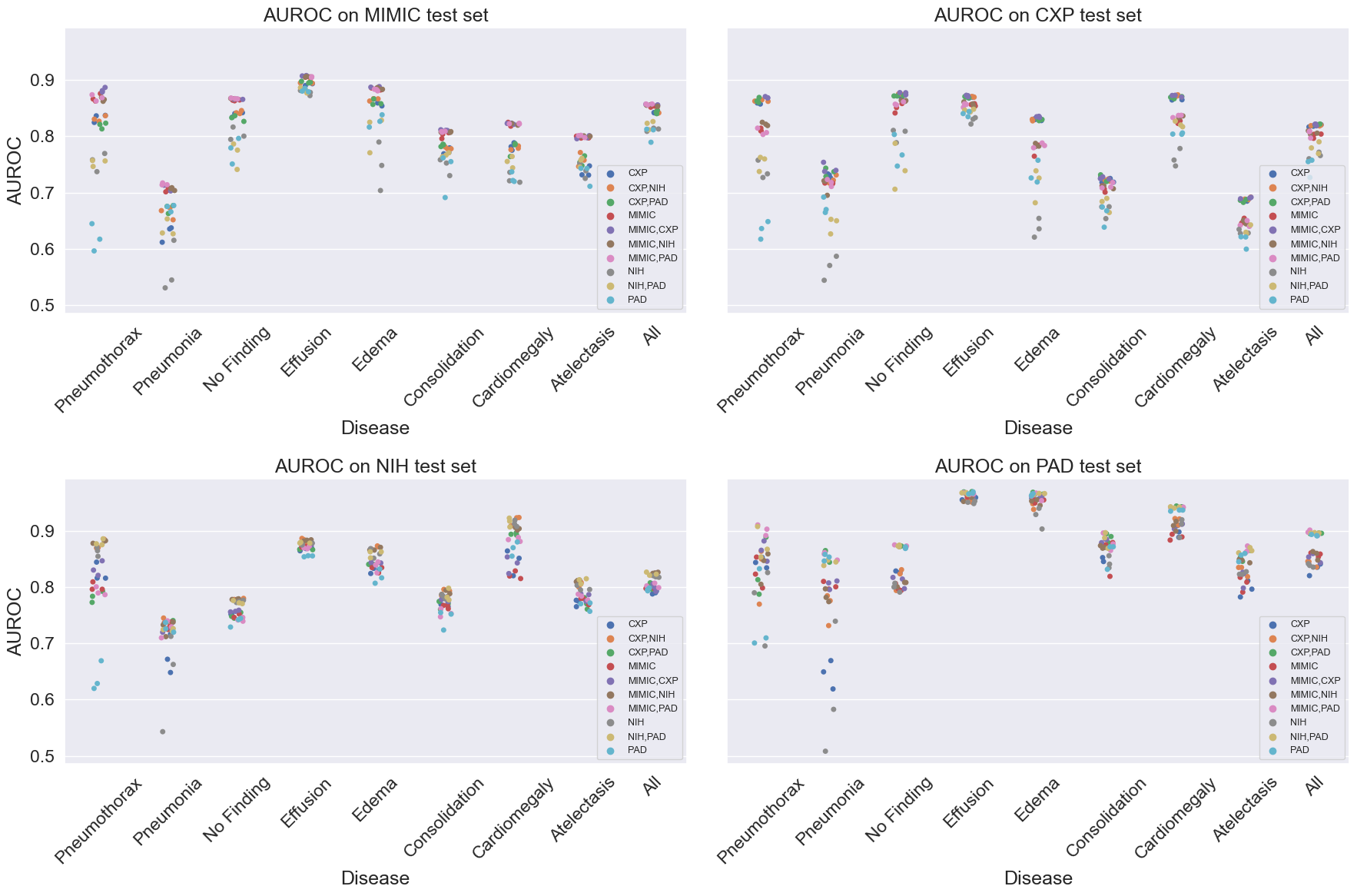}
    \caption{Absolute AUROC values for each test set's data, and across each training environment.}
    \label{fig:absolute_auroc}
\end{figure}

For completeness, we also show absolute AUROC values in \Cref{fig:absolute_auroc}, for the one-environment and two-environment models.

\subsection{Not all Datasets Have the Same Effect}

  \begin{table}[h]
\centering
\caption{Not all datasets have the same effect. Proportion of tasks (out of 27 base dataset / disease combinations) where worst-group accuracy decreases, for each dataset added.}
\label{tab:prop_of_worst_group_decr}
\begin{tabular}{l|ll}
\toprule
Added Environment & Worst Group Decreases & Balanced Worst Group Decreases \\
\midrule
CXP & 0.37 & 0.30 \\
MIMIC & 0.33 & 0.11 \\
NIH & 0.59 & 0.22 \\
PAD & 0.44 & 0.15 \\
\bottomrule
\end{tabular}
\end{table}

\Cref{tab:prop_of_worst_group_decr} shows the proportion of base environment / disease tasks where worst-group accuracy decreases when a given dataset is added, both before balancing (left) and after balancing (right). When including the datasets as-is, every dataset causes a drop in worst-group accuracy \textit{at least} 33\% of the time, with \verb|NIH| causing detriment the most often at \textasciitilde60\% of the cases. We see an improvement after balancing, with datasets being detrimental less often (e.g., \verb|NIH| causing drops goes from 59\% down to 22\%), but even after balancing, \textit{all datasets} cause a drop in worst-group performance at least 10\% of the time. Although there is variability in a dataset's benefit to performance, the detriment to generalization is not restricted to one of the chest x-ray datasets examined; this suggests the problem will be faced across many other dataset combinations.

\subsection{Embedding Analysis}

\begin{figure}[h]

\begin{subfigure}{}
\centering
\includegraphics[width=0.8\linewidth]{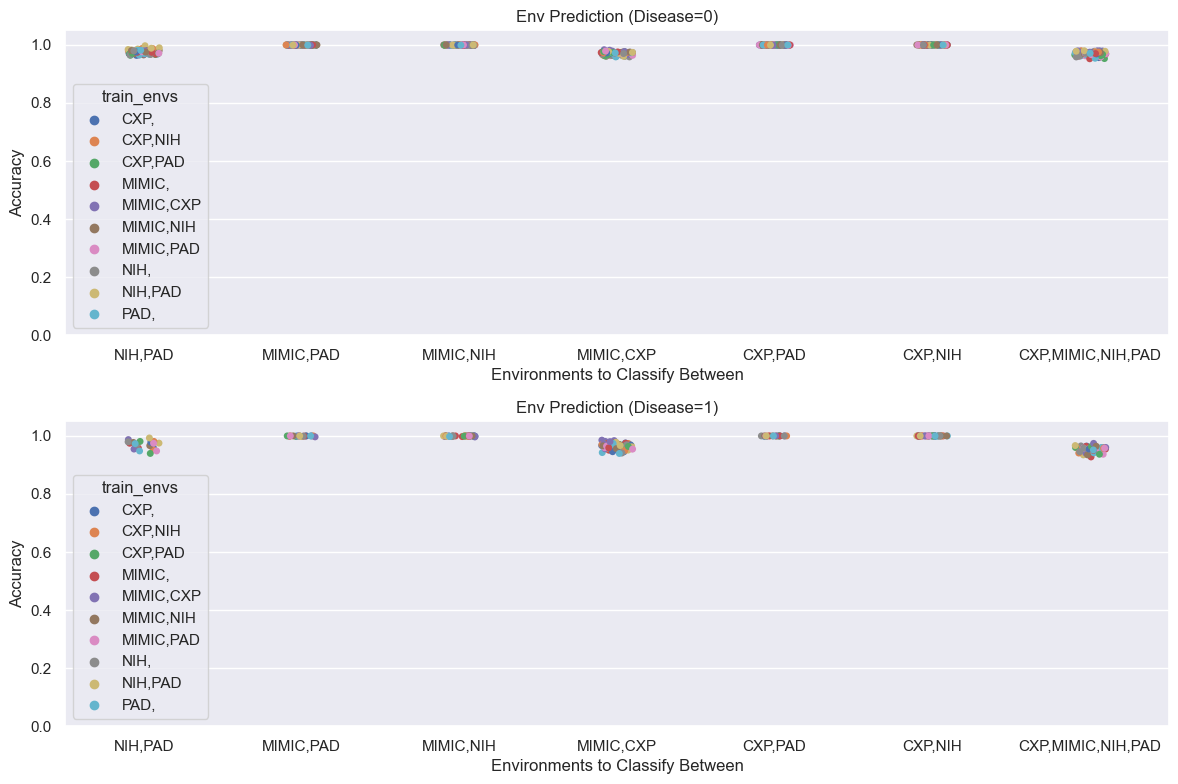}
\caption{Environment Prediction}
\label{fig:embedding-analysis-env}
\end{subfigure}


\begin{subfigure}{}
\centering
\includegraphics[width=0.6\linewidth]{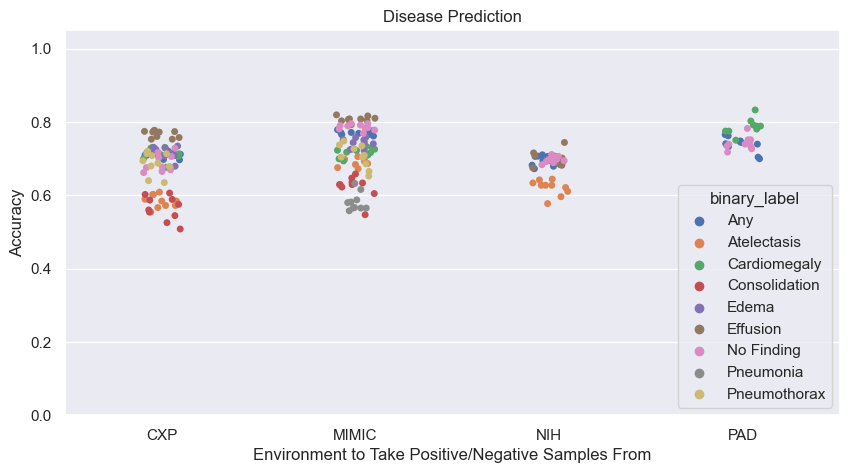}
\caption{Disease Prediction}
\label{fig:embedding-analysis-disease}
\end{subfigure}

\caption{Linear SVM validation accuracy on the two tasks we employ in our embedding analysis. Both scatter plots includes results for all nine target labels and 10 train environment configurations. The hospital prediction task can be done with near perfect accuracy, while disease prediction (the task for which the model was trained on) is only \textasciitilde70-80\% accurate.}
\label{fig:embedding-analysis-results}

\end{figure}

\Cref{fig:embedding-analysis-results} shows the results of the SVM performance for environment prediction (top) and disease prediction (bottom). The embeddings are highly discriminative for hospital, even by models that were only trained on a single hospital's data.

\subsection{Performance Decrease Summary}

\begin{figure}[h]
    \centering
    \vspace{-1.5em}
    \includegraphics[width=0.83\linewidth]{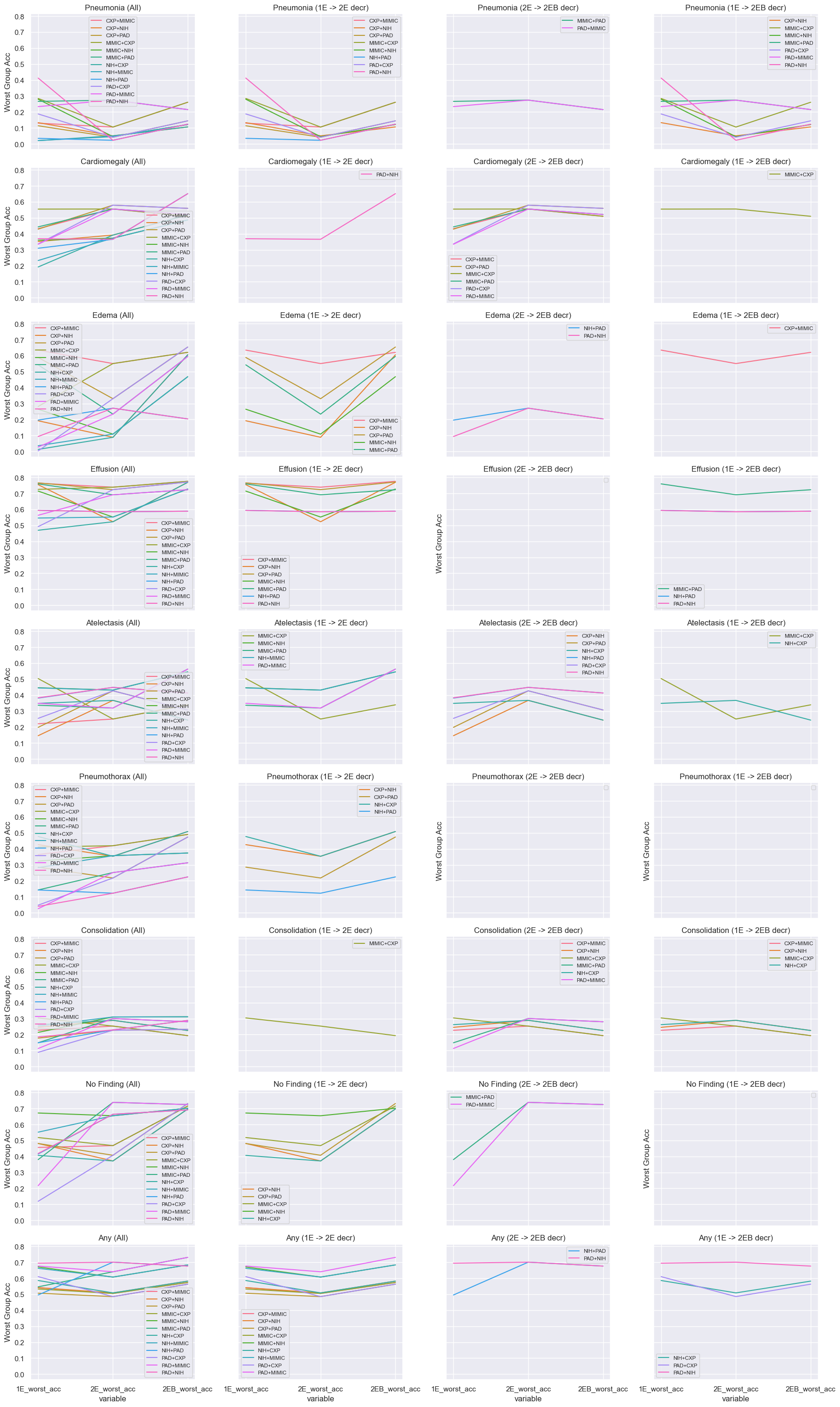}
    \caption{Performance difference between experiments}
    \label{fig:performance_decr_summary}
\end{figure}

\Cref{fig:performance_decr_summary} shows performance between different experiments. The x-axis denotes worst group accuracy when training on a single environment (left), two environments (middle), and two environments balanced (right). The columns show these same results but filtered to only show decreases in performances between each change in training setup; one hospital to two-hospital (middle left), two hospital to two hospital balanced (middle right), and one hospital to two hospital balanced (right).

\subsection{SVM Coefficients}

Given the incredible accuracy of environment prediction, a reasonable question is to ask whether this hospital-specific signal is reflected in a small number of features; this would explain the performance as they could be predicted easily in such high dimensional space.

\begin{figure}[h]
    \centering
    \includegraphics[width=0.8\linewidth]{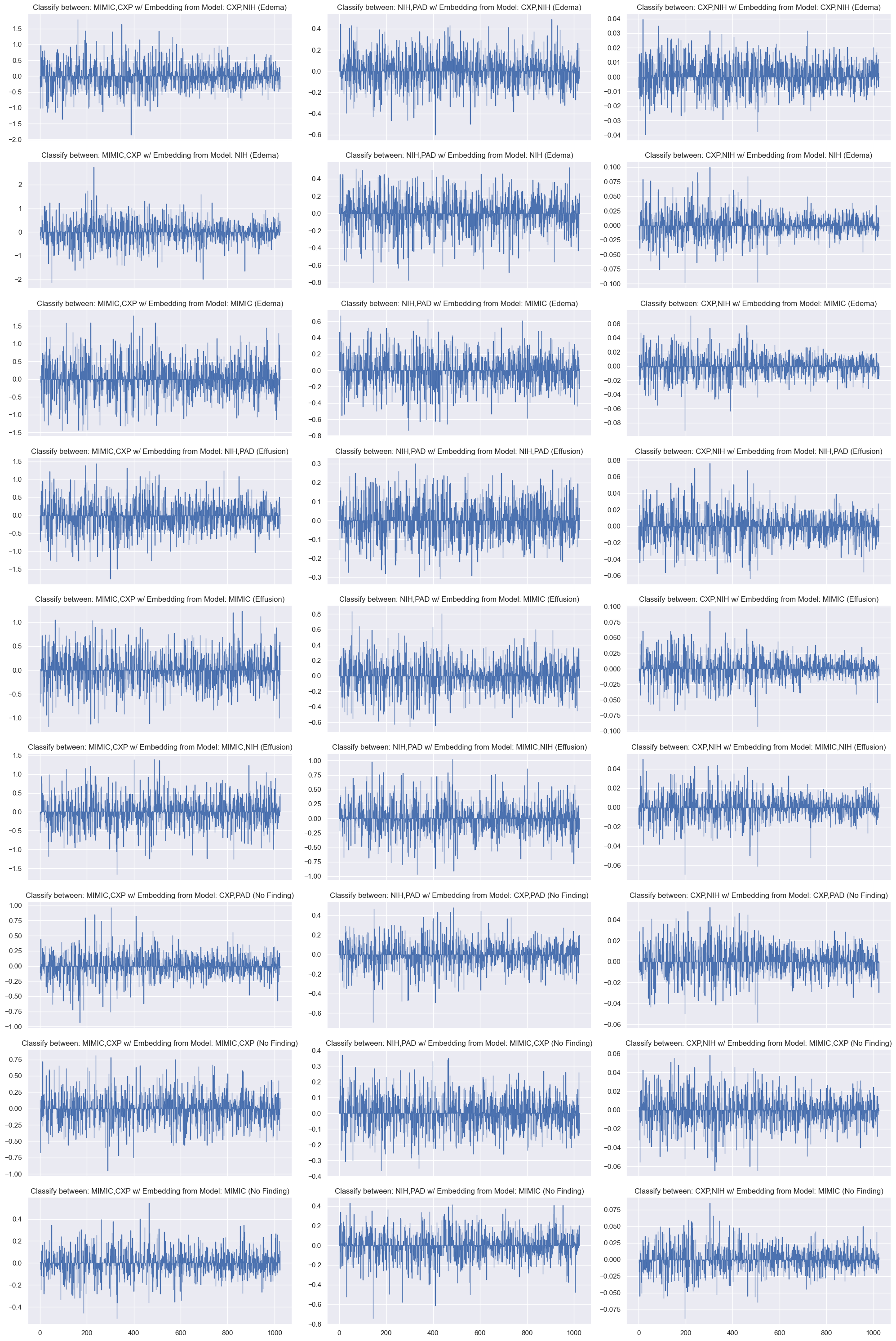}
    \caption{Linear SVM Coefficients for different embedding-based hospital classification subtasks.}
    \label{fig:svm-coefficients}
\end{figure}

In search of any instance where this phenomenon is learned, we randomly sample three SVM environment prediction subtasks (\verb|MIMIC/CXP|, \verb|NIH/PAD|, \verb|CXP/NIH|), three diseases (\textit{Edema, Effusion, No Finding}), and three sets of embeddings from each. We plot the SVM coefficients in Figure \ref{fig:svm-coefficients}. The coefficients are roughly uniform with no significant outliers/extremely strong values that would signal a hospital-specific feature in the embeddings. This means that the highly predictive hospital-specific attributes in these embeddings are distributed \textit{across the embedding-space}, making mitigation more difficult than if this signal was reflected in only a few features.

\subsection{Size-Controlled Worst Group Accuracies}

\begin{figure}
    \centering
    \includegraphics[width=\linewidth]{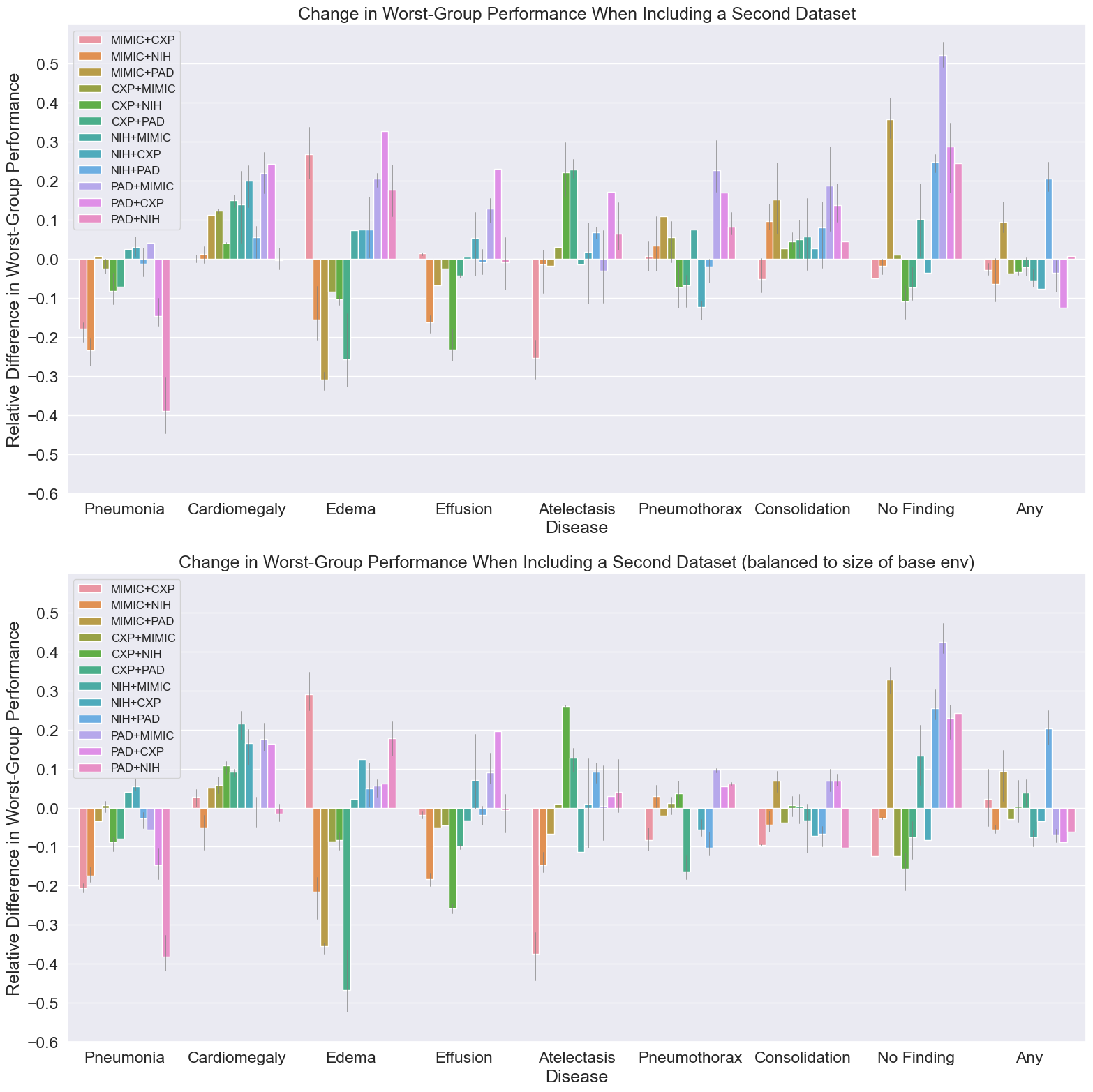}
    \caption{Change in worst-group performance after including a second training dataset. We see both improvements in performance as expected, but also many deteriorations. For every element in the legend, the first dataset is the base environment, and the second is the added environment; for instance, MIMIC+CXP shows the performance change from training on just MIMIC to training on MIMIC and CXP. The bottom figure shows the same results but the combined dataset is of the same size as the base environment (first hospital in each legend element, respectively).}
    \label{fig:1E_2E_both}
\end{figure}

\begin{figure}
    \centering
    \includegraphics[width=\linewidth]{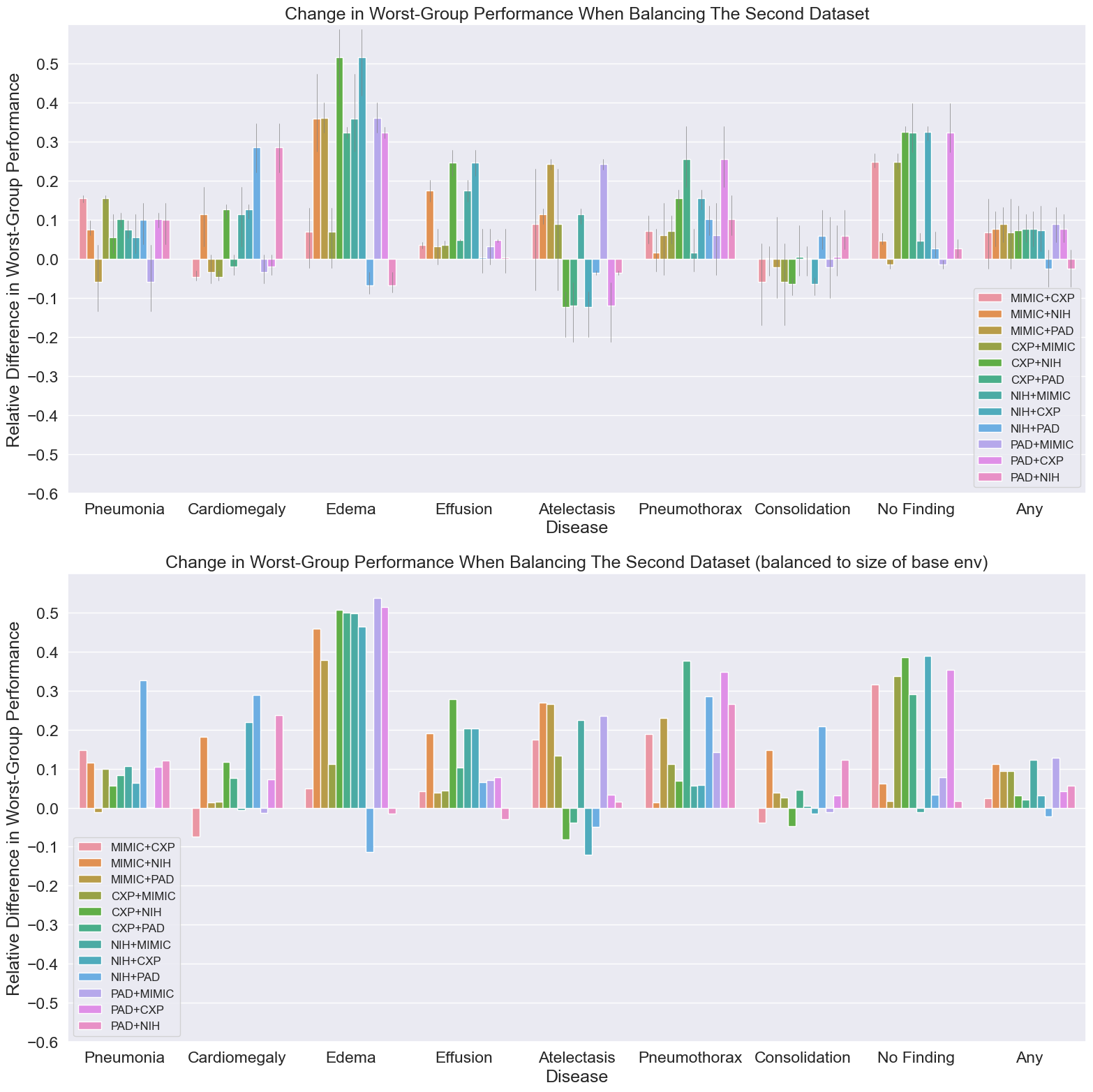}
    \caption{Change in worst-group performance when balancing the two datasets (compared to not balancing). While balancing often helps, it can also hurt performance, suggesting that it can be a promising intervention but should not be used blindly. The bottom figure shows the same results but the combined dataset is of the same size as the base environment (first hospital in each legend element, respectively).}
    \label{fig:2E_2EB_both}
\end{figure}

\begin{figure}
    \centering
    \includegraphics[width=\linewidth]{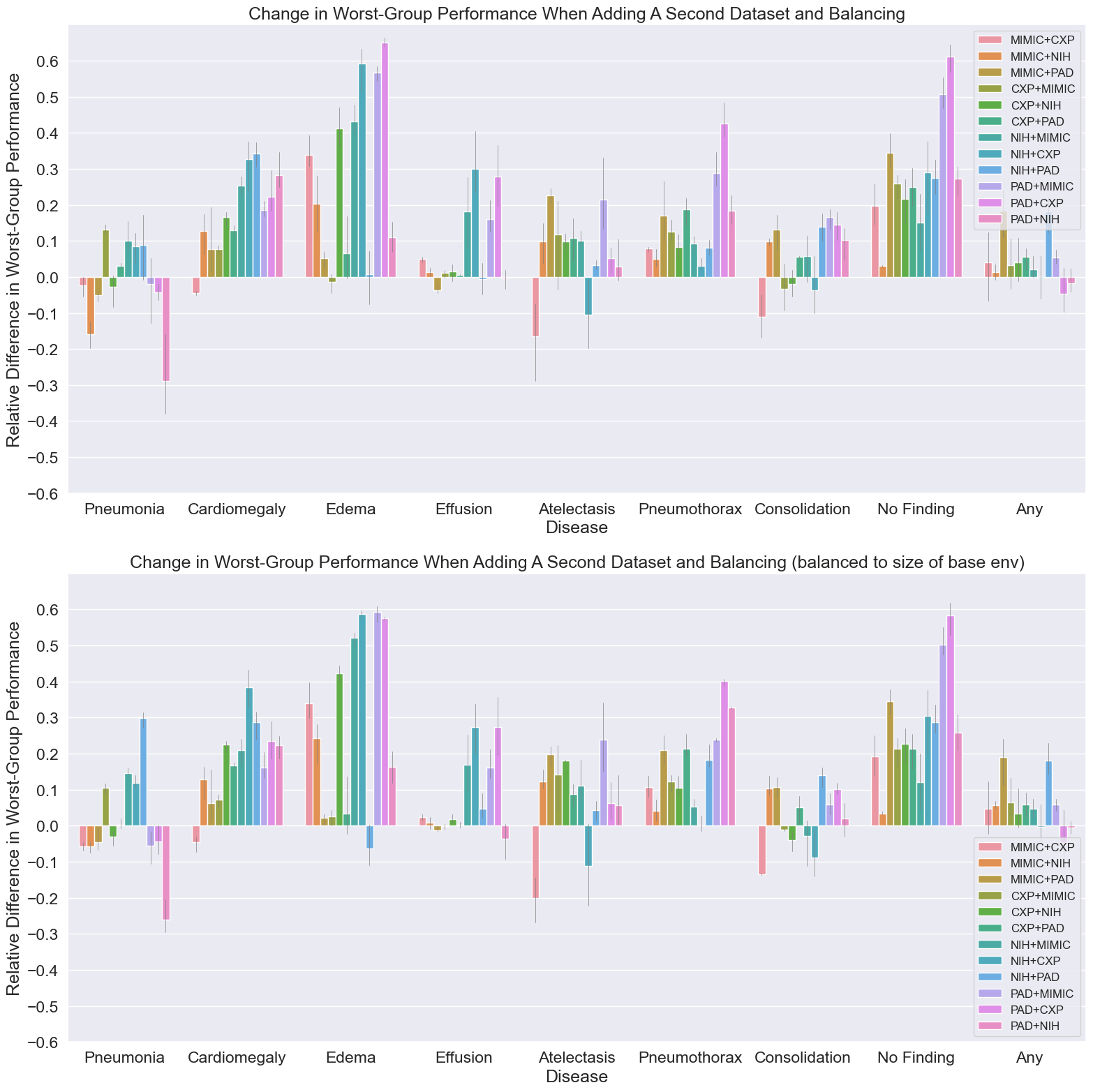}
    \caption{Change in worst-group performance after including a second dataset and balancing the two datasets. More consistently (than Figure \ref{fig:adding_second_dataset}) we see improvements but not a global increase in performance. The bottom figure shows the same results but the combined dataset is of the same size as the base environment (first hospital in each legend element, respectively).}
    \label{fig:1E_2EB_both}
\end{figure}

We show the same bar plots as above (\Cref{fig:adding_second_dataset}, \Cref{fig:app_balancing}, \Cref{fig:adding_and_balancing}) but include the size-controlled results alongside. The size controlled results are found by undersampling the combined / combined-balanced dataset to the same size as the original single source dataset, to remove dataset size as a confounder in our analysis. The results are shown below (\Cref{fig:1E_2E_both}, \Cref{fig:2E_2EB_both}, \Cref{fig:1E_2EB_both})

\end{document}